*Article*

# Your Face Mirrors Your Deepest Beliefs—Predicting Personality and Morals through Facial Emotion Recognition

**Peter A. Gloor [1], Andrea Fronzetti Colladon [2,*], Erkin Altuntas [3], Cengiz Cetinkaya [4], Maximilian F. Kaiser [5], Lukas Ripperger [5] and Tim Schaefer [5]**

1 MIT Center for Collective Intelligence, Cambridge, MA 02142, USA; pgloor@mit.edu
2 Department of Engineering, University of Perugia, 06123 Perugia, Italy
3 Galaxyadvisors AG, 5000 Aarau, Switzerland; ealtuntas@galaxyadvisors.com
4 Department of Data Science, Lucerne University of Applied Sciences and Arts, 6002 Lucerne, Switzerland; cengiz.cetinkaya@stud.hslu.ch
5 Department of Information Systems, University of Cologne, 50923 Cologne, Germany; mkaise16@smail.uni-koeln.de (M.F.K.); lripperg@smail.uni-koeln.de (L.R.); tim.schaefer94@web.de (T.S.)
* Correspondence: andrea.fronzetticolladon@unipg.it; Tel.: +39-075-585-3705

**Abstract:** Can we really "read the mind in the eyes"? Moreover, can AI assist us in this task? This paper answers these two questions by introducing a machine learning system that predicts personality characteristics of individuals on the basis of their face. It does so by tracking the emotional response of the individual's face through facial emotion recognition (FER) while watching a series of 15 short videos of different genres. To calibrate the system, we invited 85 people to watch the videos, while their emotional responses were analyzed through their facial expression. At the same time, these individuals also took four well-validated surveys of personality characteristics and moral values: the revised NEO FFI personality inventory, the Haidt moral foundations test, the Schwartz personal value system, and the domain-specific risk-taking scale (DOSPERT). We found that personality characteristics and moral values of an individual can be predicted through their emotional response to the videos as shown in their face, with an accuracy of up to 86% using gradient-boosted trees. We also found that different personality characteristics are better predicted by different videos, in other words, there is no single video that will provide accurate predictions for all personality characteristics, but it is the response to the mix of different videos that allows for accurate prediction.

**Keywords:** artificial intelligence; facial emotion recognition; personality; moral values; risk-taking; forecasting

## 1. Introduction

*A face is like the outside of a house, and most faces, like most houses, give us an idea of what we can expect to find inside. ~ Loretta Young*

*The face is the mirror of the mind, and eyes without speaking confess the secrets of the heart. ~ St. Jerome*

Proverbs like the ones above allude to the fact that our faces have the potential to give away our deepest emotions. However, just like the facade of a house might be misleading about what is inside the house, the mind behind the face might hide its true feelings. Emotionally competent people claim to be able to guess what another person is thinking by just watching that person's face. However, humans are not particularly good at reading emotions in other's faces. For instance, the test "reading the mind in the eyes" [1], which only shows the eyes of a face, is frequently answered correctly with an accuracy of less than fifty percent. Psychologist Lisa Feldmann Barrett claims that we are actually not much better than randomness when we are not primed in reading others' emotions [2]. Humans are also notoriously bad at identifying personality characteristics in others [3]. While early systems to read emotions from the face were extracting features from different parts of the face, and comparing them directly, for instance on the basis of the facial action coding system FACS [4], facial emotion recognition has made huge progress over the last 10 years thanks to advances in AI and deep learning [5–7]. In this paper, we used latest advances in this field to automatically predict personality characteristics calibrated using four well-established frameworks assessing different facets of personality: Neo-FFI [8], moral foundations [9], Schwartz moral values [10], and attitudes towards taking risk [11].

The remainder of this paper is organized as follows. First, we set the stage by explaining how the emotional response to an external event can demonstrate the moral values of an individual. We also motivate how facial expressions might indicate the personality characteristics of a person. We then introduce our system that tracks emotions through facial emotion recognition while the viewer is watching a video, with 15 small emotionally triggering video snippets. We then present our results, demonstrating through correlations, regression, and machine learning that the emotional response in the face of the viewer, captured through face emotion recognition, will indeed predict the personality and moral values of the viewer. We conclude the paper with a discussion of the results, limitations, and future work.

## 2. Background

### *2.1. Emotional Response Shows Individual Value System*

On the basis of their moral values, humans experience or show different emotions in response to an external stimulus. Emotional actions triggered through moral values are called "moral affect" [12]. Moral affect—such as shame, guilt, and embarrassment—is linked to moral behavior, leading to prohibitions against behavior that is likely to have negative consequences for the well-being of others [13]. For instance, on the basis of the personal value system, an individual might have shown a different emotional reaction when President Trump was announcing the construction of a wall to keep out asylum seekers from Mexico [14]. Both philosophers [14] and psychologists [2] have investigated this link between morals and emotions. In order to experience that something is wrong, one needs to have a feeling of disapproval towards it [14]. To measure this feeling of disapproval, thus far, technologies such as tracking the hormone level in blood or saliva have been used. For instance, it has been shown that the hormone level in saliva of homosexual and heterosexual men, when shown pictures of two men kissing, is radically different [15]. The researchers showed homosexual and heterosexual men in Utah pictures of same-sex public display of affection, plus disgusting images, such as a bucket of maggots. They used the link between disgust and prejudice, which has been shown to be capable of eliciting responses from the sympathetic nervous system, one of the body's major stress systems [16]. Salivary alpha-amylase is considered a biomarker of the sympathetic nervous system that is especially responsive to inductions of disgust. The researchers found that the difference in salivary alpha-amylase explained the degree of sexual prejudice against homosexuality among their test subjects, similar to their disgust about a bucket of maggots. In other words, their emotional response, measured through salivary alpha-amylase, indicated their moral values. Instead of measuring negative (and positive) emotions through the saliva, in our research, we measured it through face emotion recognition, maintaining the existence of a similar link between emotional response and moral values.

### *2.2. Reading Personality Attributes from Facial Characteristics*

Studying the relationship between facial and personality characteristics has a long history going back to antiquity. The book "*Physiognomics*", discussing the relationship between facial appearance and character, was written 300 BC in Aristotle's name, but is today attributed to a different author by most researchers. Swiss poet, writer, philosopher, physiognomist, and theologian Johann Caspar Lavater published between 1775 and 1778 his magnum opus on physiognomy, "*Physiognomische Fragmente zur Beförderung der Menschenkenntnis und Menschenliebe*" (Physiognomic fragments to promote knowledge of human nature and human love) [17], which cataloged leaders and ordinary men (there were very few pictures of women) of his time by their facial shape, or what he called their "lines of countenance". Lavater even invented an apparatus for taking facial silhouettes to quickly capture the characteristics of a face, and thus the personality of the person.

Later, statistician Francis Galton tried to define physiognomic characteristics of health, beauty, and criminology by creating composites through overlaying pictures of archetypical faces [18]. Italian criminologist and scientist Cesare Lombroso continued this work by defining facial measures of degeneracy and insanity including facial angles, "abnormalities" in bone structure, and volumes of brain fluid [19]. For the better part of the 20th century, scientists derogatively titled physiognomics as "pseudoscience". This changed towards the end of the 20th century. While early physiognomists from Aristotle to Lombroso tried to develop manually assembled frameworks, AI and deep learning has given a huge boost to this emerging field. Recently, physiognomics has been experiencing renewed interest by researchers, particularly by comparing facial width to height ratio with personality. The theory of "facial width to height ratio" (fWHR) posits that men with higher "facial width to height ratio", that means with broader, rounder faces, are more aggressive, while men with thinner faces are more trustworthy [20–23].

Recognizing these features automatically through facial emotion recognition has come a long way since the early days of the facial action coding system, thanks to recent advances in AI and deep learning. A large amount of research has addressed the issue of recognizing personality characteristics from facial attributes. For instance, ChaLearn "Looking at People First Impression Challenge" released a dataset with 10,000 15 s videos with faces (https://chalearnlap.cvc.uab.cat/dataset/20/description/, accessed on 21 December 2021) [24], asking participants in the challenge to identify the FFI personality characteristics [8] of the person on the video, and their age, ethnicity, and gender attributes [25]. The problem with this dataset is that the personality attributes had been added by Amazon Mechanical Turkers, which sometimes leads to a biased ground truth, as it is based on guesswork by humans (the turkers). As was mentioned in the introduction, it has been shown by other researchers that accuracy of human labelers in recognizing emotions is only incrementally better than guesswork at slightly below 50 percent [2]. Nevertheless, the winners of the ChaLearn challenge have achieved impressive accuracy on this pre-labeled dataset to correctly predict the FFI personality characteristics at over 91% [26]. However, it would be better to have true ground truth on the personality characteristics of the subjects on the video. In another project using Facebook likes, where ground truth was available, the researchers showed that the computer was actually better in recognizing personality characteristics than work colleagues, who reached only 27% accuracy, while the computer achieved 56% accuracy [3]; spouses were the most accurate at 58%. The personality characteristics had been collected from 86,220 users through a personality survey on Facebook and were predicted through Facebook likes using regression.

Earlier work has used facial expression of the viewer to measure the quality of a video [27–29]. We extend this work to not only measure the degree of enjoyment of the viewer, but the personality characteristics and moral values of the viewer—motivated by the insight that facial expressions will mirror moral values—combining face emotion recognition with ground truth obtained directly from surveys taken by the individual.

## 3. Methodology—Recording Emotions While Watching Videos

Our approach extends existing systems by not only measuring video quality, but moral values and personality of the viewers, as it uses real ground truth on personality characteristics and moral values for prediction by asking the people whose faces are recorded while watching a sequence of 15 emotionally touching video segments to also fill out a series of personality characteristics tests.

### *3.1. Measuring Facial Emotions*

Our system consists of a website (coinproject.compel.ch, accessed on 21 December 2021) where the participant watches a sequence of 15 videos (Figure 1).

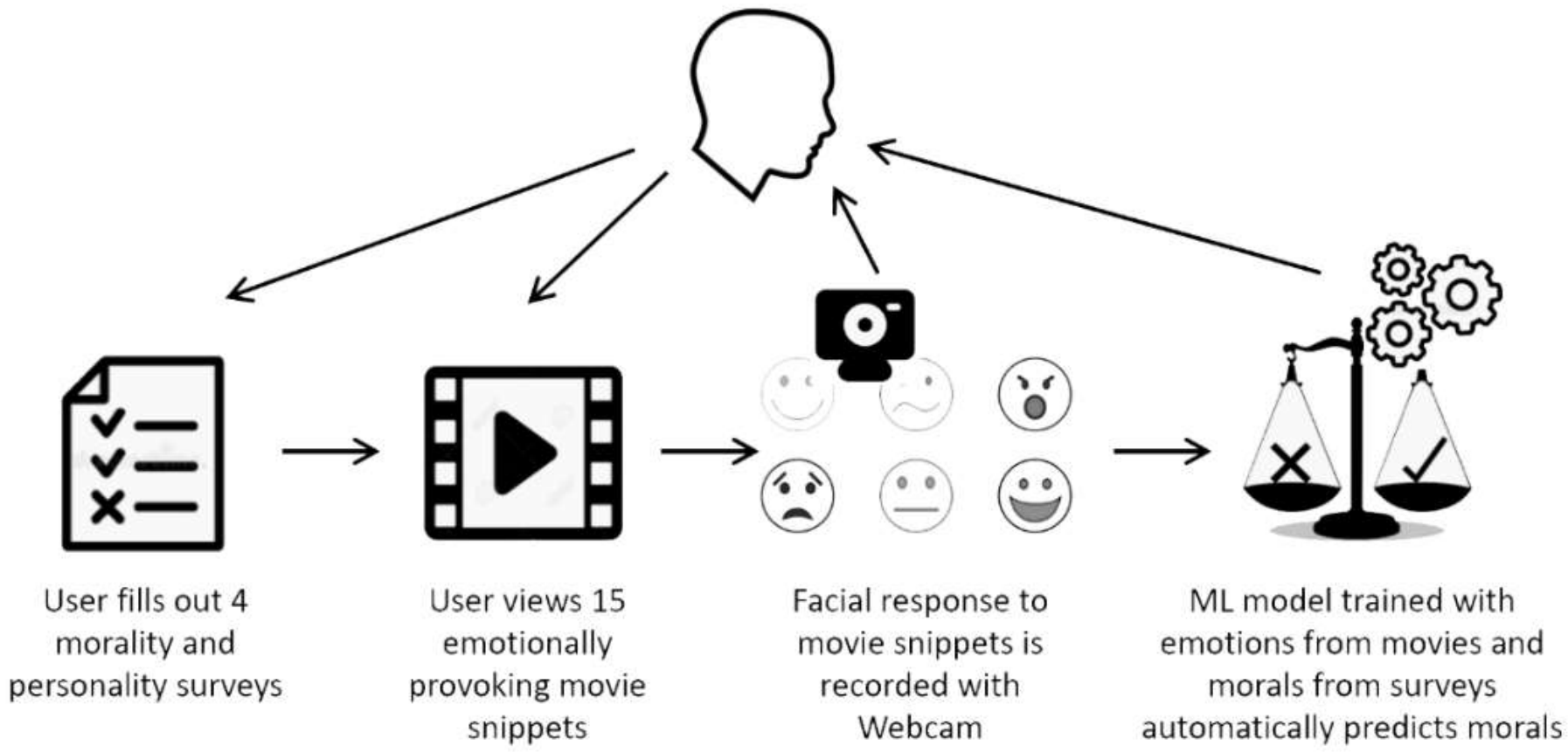

**Figure 1.** Setup of our system with video website and four online surveys.

Table 1 lists the 15 movie snippets, at a total length of 9 min 22 s, that are shown to users on the website, while the emotions of their faces are recorded after they have given informed consent that their anonymized emotions will be recorded; no video of the face is recorded.

**Table 1.** List of 15 movie snippets.

| Video Number | Short Description |
|---|---|
| 1 | puppies—cute puppies running |
| 2 | avocado—a toddler holding an avocado |
| 3 | condom ad—child throwing a tantrum in a supermarket |
| 4 | runner—competitive runners supporting a girl from another team over the finish line |
| 5 | maggot—a guy eating a maggot |
| 6 | soldier—soldiers at battle |
| 7 | Trump—Donald Trump talking about the Mexican mass migration |
| 8 | mountain bike—mountain biker on daring ride down a rock bridge |
| 9 | roof bike—guy biking on top of a skyscraper |
| 10 | roof run—guy balancing and almost falling on top of skyscraper |
| 11 | racoon—man beating racoon to death |
| 12 | abandoned—social worker feeding a starved abandoned black toddler |
| 13 | waste—residents collecting electronic waste in the slums of Accra |
| 14 | dog—sad dog on the gravestone of his master, missing him |
| 15 | monster—man discovering an invisible monster through the picture on his instant camera |

The 15 video snippets show controversial scenes with the aim of generating a wide range of emotions in respondents [30]. We use the face-api.js tool (https://justadudewhohacks.github.io/face-api.js/docs/index.html, accessed on 21 December 2021), which employs a convolutional neural network with a ResNet-34 architecture [31], to recognize the user's facial emotions in each frame (up to 30 times per second) of the user's web cam. The tracked emotions are joy, sadness, anger, fear, surprise, and disgust [32]. In addition, a seventh emotion "neutral" was added, which greatly increases machine learning accuracy when none of the six Ekman emotions can be recognized.

### *3.2. Measuring Personality and Morals of the Viewers*

Our dependent variables are collected through four well-validated personality and moral values assessments. The user is asked on the same website where the videos are shown to fill out four online surveys for the revised NEO FFI personality inventory, the Haidt moral foundations test, the Schwartz personal value system, and the domain-specific risk-taking scale (DOSPERT). The OCEAN (Openness, Conscientiousness, Extroversion, Agreeability, Neuroticism) personality characteristics are measured with the Neo-FFI [8] survey. Risk-preference is measured by the Domain-Specific Risk-Taking (DOSPERT) survey [11], which assesses disposition to take risks in five specific domains of life (ethical, financial, recreational, health, and social). It measures both the willingness to take risks and the individual perception of an activity as risky. Moral foundational values are measured with the Haidt moral foundations survey [9]. It measures the moral values of the respondent in five categories (care, fairness, loyalty, authority, and sanctity). In addition, the two dimensions of Conservation and Transcendence also are assessed through a survey [10,33]. The Schwartz values have been validated in many countries around the world [34].

## 4. Results—Emotional Response Predicted Values

We found that all four dimension of a personality, FFI characteristics, DOSPERT risk taking, moral foundations, and Conservation and Transcendence (Schwartz values), can be predicted on the basis of the emotions shown while watching the 15 different video segments. Table 2 shows the descriptive statistics of our dependent variables for all four dimensions of a personality, listing the individual traits we mapped through psychometric tests.

**Table 2.** Descriptive statistics of individual traits.

| Variable | M | SD | Min | Max |
|---|---|---|---|---|
| Agreeableness | 0.64 | 0.08 | 0.47 | 0.83 |
| Conscientiousness | 0.69 | 0.06 | 0.52 | 0.83 |
| Neuroticism | 0.54 | 0.09 | 0.33 | 0.73 |
| Extraversion | 0.67 | 0.07 | 0.50 | 0.83 |
| Openness to experience | 0.61 | 0.06 | 0.48 | 0.78 |
| ETH_L | 2.56 | 1.31 | 1.50 | 7.33 |
| ETH_P | 4.55 | 1.23 | 1.83 | 8.83 |
| FIN_L | 3.25 | 1.39 | 1.00 | 8.33 |
| FIN_P | 4.72 | 1.34 | 1 | 9 |
| HEA_L | 3.33 | 1.10 | 1.17 | 6.33 |
| HEA_P | 4.81 | 1.02 | 1.50 | 7.17 |
| SOC_L | 5.58 | 1.07 | 3.50 | 9.67 |
| SOC_P | 2.72 | 1.12 | 1.17 | 6.67 |
| REC_L | 4.19 | 1.35 | 1.50 | 7.33 |
| REC_P | 3.99 | 1.15 | 1.83 | 7 |
| Conservation | 0.77 | 0.74 | −0.62 | 3.54 |
| Transcendence | −1.20 | 0.70 | −2.87 | 0.70 |
| Harm/care | 22.36 | 3.93 | 12 | 29 |
| Fairness/reciprocity | 22.13 | 4.08 | 7 | 30 |
| In-group loyalty | 16.54 | 4.30 | 6 | 25 |
| Authority/respect | 13.57 | 4.45 | 3 | 22 |
| Purity/sanctity | 11.93 | 4.68 | 0 | 20 |

ETH_L = ethical likelihood; ETH_P = ethical perceived; FIN_L = financial likelihood; FIN_P = financial perceived; HEA_L = health likelihood; HEA_P = health perceived; SOC_L = social likelihood; SOC_P = social perceived; REC_L = recreational likelihood; REC_P = recreational perceived.

In Appendix A, we show the Pearson's correlation coefficients of individual traits with the different emotions experienced while watching the videos. Neither commenting on each single association and its significance, nor investigating the possible reasons behind associations, is in the scope of this research. Rather we wanted to show the possibility of predicting individual traits, based on the differential emotional response of individuals exposed to the same set of stimuli, by considering automatically recognized emotions through artificial intelligence.

The preliminary result of correlations—a suggested association between individual differences and people's emotional responses—is confirmed by the regression models presented in Tables 3–6. For each set of dependent variables, they show the best model, i.e., the optimal combination of predictors that can explain the larger proportion of variance. We found no evidence of collinearity problems (evaluated by calculating variance inflation factors). These regressions illustrate the predictability of personality characteristics and morals from facial expression of emotions using conventional statistical methods.

**Table 3.** Regression models for the Big Five personality traits.

| Predictor/Dependent | Neuroticism | Extraversion | Openness to Experience | Agreeableness | Consciousness |
|---|---|---|---|---|---|
| Angry 2 | | | 4.922 *** | | |
| Angry 7 | 0.665 * | | | | |
| Disgusted 4 | | 0.297 * | | | |
| Disgusted 11 | | | | 0.547 ** | |
| Happy 1 | | | | | −0.067 ** |
| Happy 8 | | 0.070* | | | 0.149 *** |
| Happy 9 | | | | | −0.104 ** |
| Happy 13 | −0.398 * | | | | |
| Happy 15 | 0.226 ** | | | | |
| Neutral 1 | | | 0.038 * | | |
| Neutral 10 | | | | 0.062 * | |
| Surprised 7 | | | −2.735 ** | | |
| Surprised 9 | | 0.499 ** | | | |
| Surprised 11 | | | 0.352 * | | |
| Surprised 14 | | −3.049 ** | | | |
| Constant | 0.518 *** | 0.659 *** | 0.582 *** | 0.591 *** | 0.700 *** |
| Adjusted $R^2$ | 0.184 | 0.263 | 0.257 | 0.131 | 0.198 |
| N | 80 | 80 | 80 | 80 | 80 |

* $p < 0.05$; ** $p < 0.01$; *** $p < 0.001$.

In general, we found that models for some traits—such as conservation, transcendence, and ethical and financial likelihood—had promising adjusted R2 values. In terms of emotions, fear seemed more relevant for the predictions of the DOSPERT scores, whereas happiness seemed more associated with the Big Five personality traits. Being neutral in front of a video can also play a role in determining the individual's personality characteristics. Remember that the facial emotion recognition system returns this value if it cannot assign any other emotion with a sufficiently high threshold, corresponding to the individual sitting in front of the computer with an unmoving face. We also see that different videos triggered a variety of emotional responses, which were possibly useful for the prediction of different traits. All the relationships explored in this study could be further investigated in future research in order to better analyze their meaning from a psychological perspective.

Figure 2 summarizes findings from the regression models, providing evidence to the importance of each video and emotion for the prediction of individual traits.

For example, we can observe that videos number 14, 9, and 2 were those that triggered the most useful emotional responses. Among emotions, fear and happiness were those most used to make predictions, with fear being particularly relevant for the DOSPERT traits.

*Predicting Personality and Morals Using Machine Learning*

While correlations and regressions showed promising results, we wanted to complete our analysis to explore non-linear relationships and the possibility of making predictions by using machining learning and considering a test sample (a subset of observations) not used for model training. In particular, we binned the continuous scores of our dependent variables into three classes in order to understand if values were high, medium, or low. Subsequently, we used a gradient boosting approach to make predictions, namely, Xgboost [35]. We trained our models using 10-Fold Cross Validation and the SMOTE technique [36] in order to treat unbalanced classes. ADASYN was also used as an alternative to SMOTE [37], in the cases where this led to improved forecasts. In Table 7, we present the results of these forecasting exercises, made on 10% of observations that were held out for testing prediction accuracy.

**Table 4.** Regression models for the DOSPERT scale values.

| Predictor/Dependent | ETH_L | ETH_P | FIN_L | FIN_P | HEA_L | HEA_P | SOC_L | SOC_P | REC_L | REC_P |
|---|---|---|---|---|---|---|---|---|---|---|
| Angry 1 | −70.349 *** | | | | | | | | | |
| Angry 4 | | | | | | | | 22.442 ** | | |
| Angry 5 | 35.700 *** | | | | | | | | | |
| Disgusted 2 | | −59.387 * | | | | | | | | |
| Disgusted 15 | | 62.642 ** | | | | | | | | |
| Fearful 1 | | | | | −146.951 * | | | | | |
| Fearful 3 | | | | −119.941 * | | | | | | −103.950 * |
| Fearful 4 | | | | | | | 125.975 *** | | | |
| Fearful 7 | | 71.613 *** | | | 67.482 ** | 101.615 *** | | | | |
| Fearful 8 | | | −150.002 * | | | | | | | |
| Fearful 9 | 20.145 *** | | 32.366 *** | | | | | | 13.496 * | 11.481 * |
| Fearful 10 | | | | 105.252 *** | | | | | | |
| Fearful 13 | | | | | | | | | | 180.138 ** |
| Happy 8 | | | | | | 1.301 ** | | | | |
| Happy 12 | | | | | | | | | 12.194 * | |
| Happy 14 | | | | | | | | | −5.893 ** | |
| Neutral 5 | | 1.032 * | | | | | | | | |
| Neutral 6 | | | | | | | | | | −1.107 ** |
| Sad 2 | | | 3.487 * | | | | | | | |
| Sad 5 | | | −3.118 *** | | | | | | | |
| Sad 15 | | | | | | | −1.014 * | | | |
| Surprised 5 | | | | | | −19.917 ** | | | | |
| Surprised 6 | | −16.358 ** | | | | | | | | |
| Constant | 2.308 *** | 4.022 *** | 3.721 *** | 4.696 *** | 3.333 *** | 4.682 *** | 5.615 *** | 2.613 *** | 4.211 *** | 4.692 *** |
| Adjusted $R^2$ | 0.370 | 0.318 | 0.389 | 0.190 | 0.188 | 0.276 | 0.245 | 0.090 | 0.177 | 0.273 |
| N | 65 | 65 | 65 | 65 | 65 | 65 | 65 | 65 | 65 | 65 |

* $p < 0.05$; ** $p < 0.01$; *** $p < 0.001$. ETH_L = ethical likelihood; ETH_P = ethical perceived; FIN_L = financial likelihood; FIN_P = financial perceived; HEA_L = health likelihood; HEA_P = health perceived; SOC_L = social likelihood; SOC_P = social perceived; REC_L = recreational likelihood; REC_P = recreational perceived.

**Table 5.** Regression models for conservation and transcendence.

| Predictor/Dependent | Conservation | Transcendence |
|---|---|---|
| Happy 4 | | 1.078 ** |
| Happy 5 | −0.780 * | |
| Happy 8 | | −1.527 *** |
| Happy 10 | −0.829 ** | |
| Fearful 14 | −13.963 * | −15.090 ** |
| Surprised 2 | | 26.934 * |
| Surprised 14 | 44.428 *** | |
| Constant | 0.952 *** | −1.232 *** |
| Adjusted $R^2$ | 0.341 | 0.280 |
| N | 70 | 70 |

* $p < 0.05$; ** $p < 0.01$; *** $p < 0.001$.

**Table 6.** Regression models for the Haidt moral values.

| Predictor/Dependent | Harm/Care | Fairness/Reciprocity | In-Group Loyalty | Authority/Respect | Purity/Sanctity |
|---|---|---|---|---|---|
| Angry 4 | | | | 71.456 * | |
| Happy 3 | −4.152 * | | −3.442 * | | |
| Happy 10 | | | | −5.078 ** | |
| Neutral 2 | −7.783 *** | | | | |
| Neutral 6 | | −7.897 *** | | | |
| Neutral 7 | | | −5.329 ** | | |
| Neutral 10 | | 3.391 * | | | |
| Sad 2 | | | | | 16.123 ** |
| Surprised 14 | | | | 143.987 * | 179.091 ** |
| Constant | 28.435 *** | 25.585 *** | 21.432 *** | 13.294 *** | 10.433 *** |
| Adjusted $R^2$ | 0.261 | 0.193 | 0.115 | 0.197 | 0.200 |
| N | 69 | 69 | 69 | 69 | 69 |

* $p < 0.05$; ** $p < 0.01$; *** $p < 0.001$.

**Table 7.** Accuracy of Xgboost models.

| Variable | Average Accuracy | Cohen's Kappa |
|---|---|---|
| Conservation | 73.3% | 0.57 |
| Transcendence | 71.4% | 0.53 |
| Authority/respect | 73.3% | 0.61 |
| Fairness/reciprocity | 73.3% | 0.56 |
| Harm/care | 86.7% | 0.79 |
| In-group loyalty | 80.0% | 0.66 |
| Purity/sanctity | 73.3% | 0.58 |
| Agreeableness | 81.3% | 0.71 |
| Conscientiousness | 78.9% | 0.69 |
| Extraversion | 72.2% | 0.58 |
| Neuroticism | 82.3% | 0.73 |
| Openness to experience | 72.2% | 0.58 |
| Ethical likelihood | 78.6% | 0.65 |
| Ethical perceived | 78.6% | 0.68 |
| Financial likelihood | 84.6% | 0.77 |
| Financial perceived | 78.6% | 0.68 |
| Health likelihood | 84.6% | 0.75 |
| Health perceived | 60.0% | 0.38 |
| Recreational likelihood | 71.4% | 0.59 |
| Recreational perceived | 86.7% | 0.80 |
| Social likelihood | 76.9% | 0.63 |
| Social perceived | 71.4% | 0.57 |

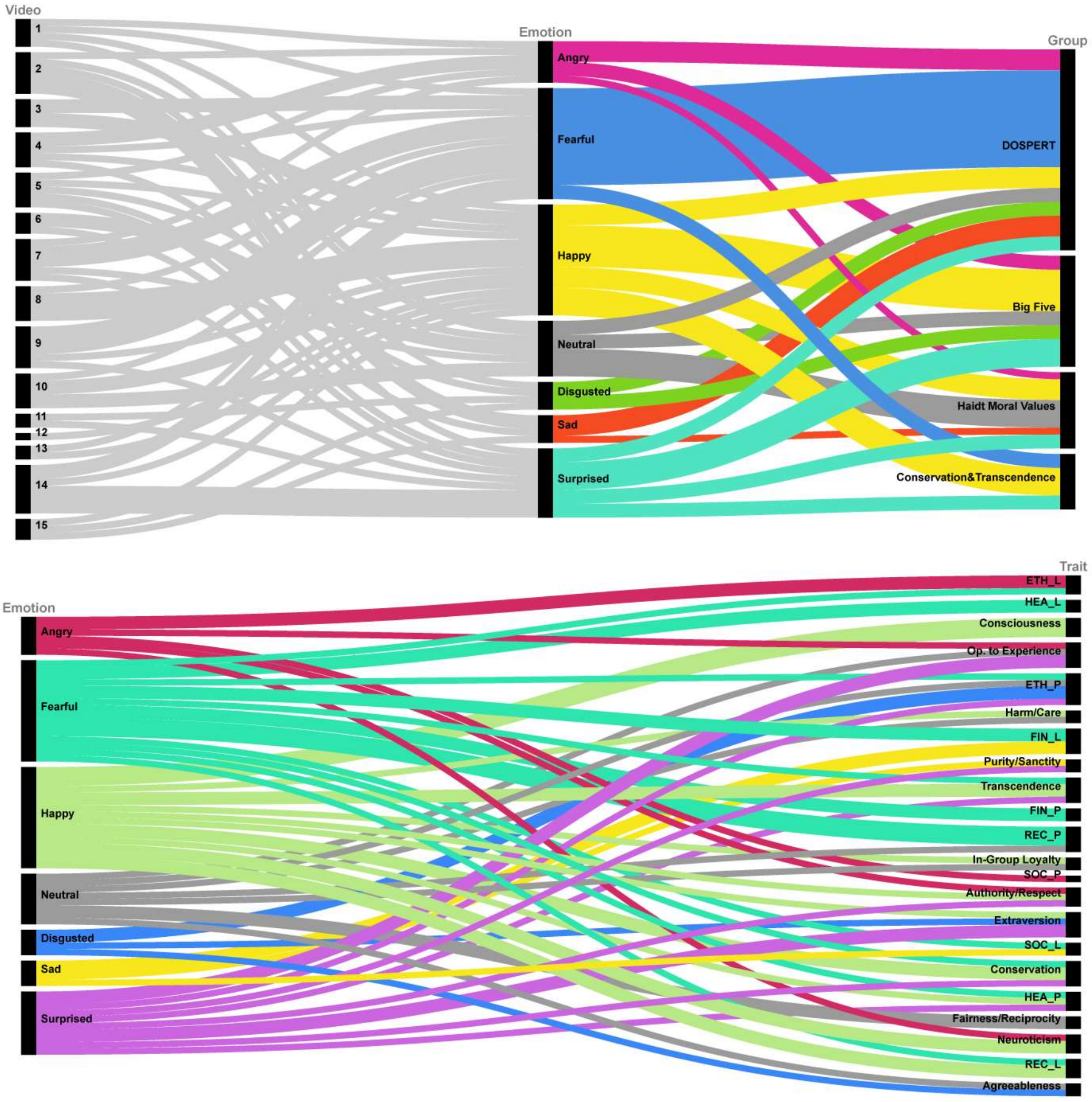

**Figure 2.** Alluvial diagrams illustrating the significant relationships between videos, emotions, and personality (**top**), and between emotions and individual traits (**bottom**).

As the table shows, we obtained good prediction results, both in terms of average accuracy and Cohen's Kappa. Only for the health perceived trait of the DOSPERT scale did we obtain an accuracy score that was below 70% (60% average accuracy and a Kappa value of 0.38). This confirms our original hypothesis that facial emotion recognition can be used to predict personality and other individual traits.

Similarly to the regression models, different features were more important for the prediction of personality and other individual traits. In order to evaluate the contribution of each feature to model prediction, we used Shapley additive explanations (SHAP) [38,39]. In the following (Figures 3–6), we provide some examples, while the remaining charts are shown in Appendix A.

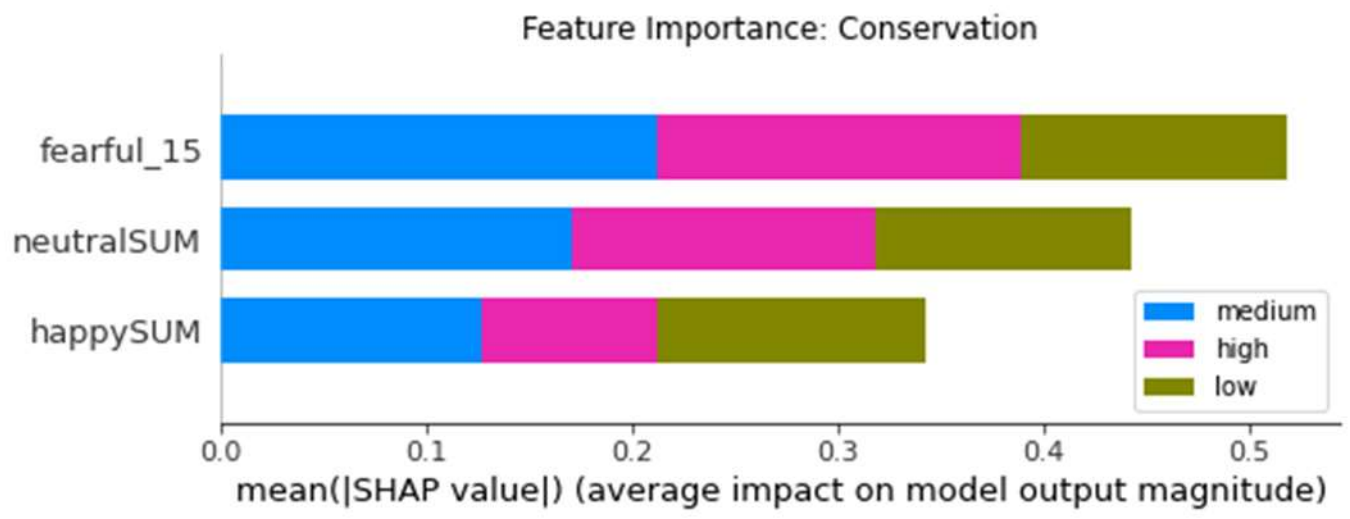

**Figure 3.** Feature importance for predicting conservation.

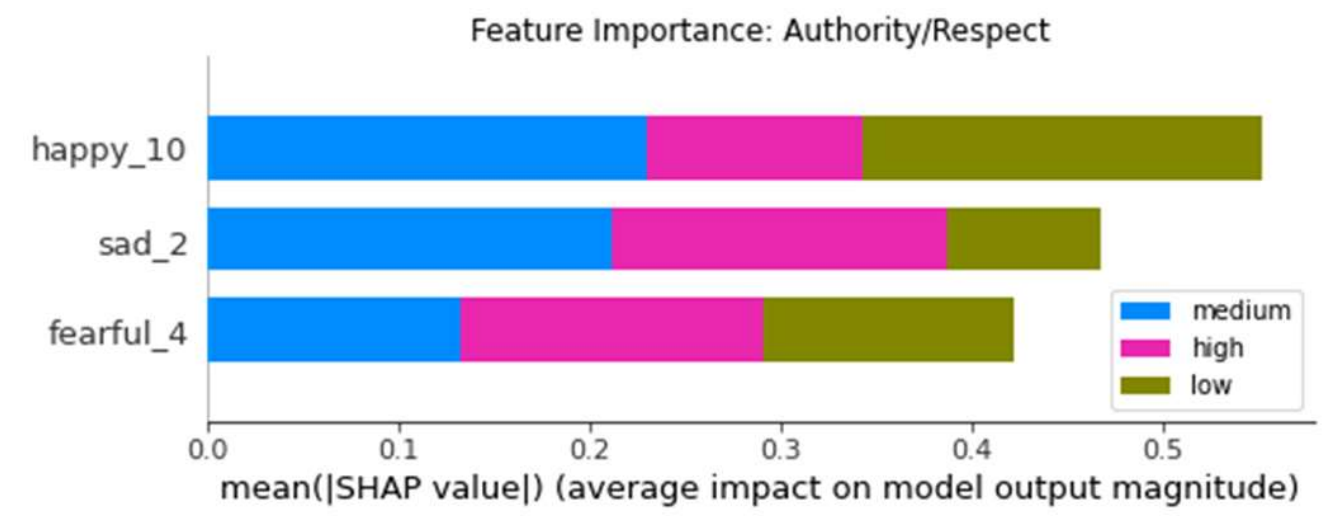

**Figure 4.** Feature importance for predicting authority/respect.

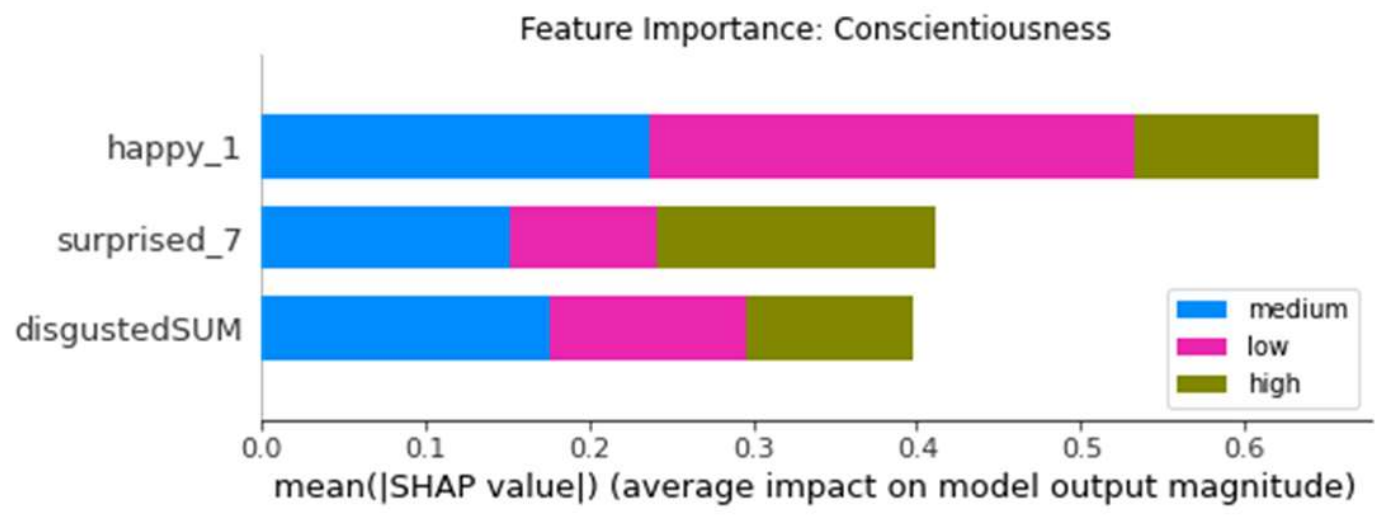

**Figure 5.** Feature importance for predicting conscientiousness.

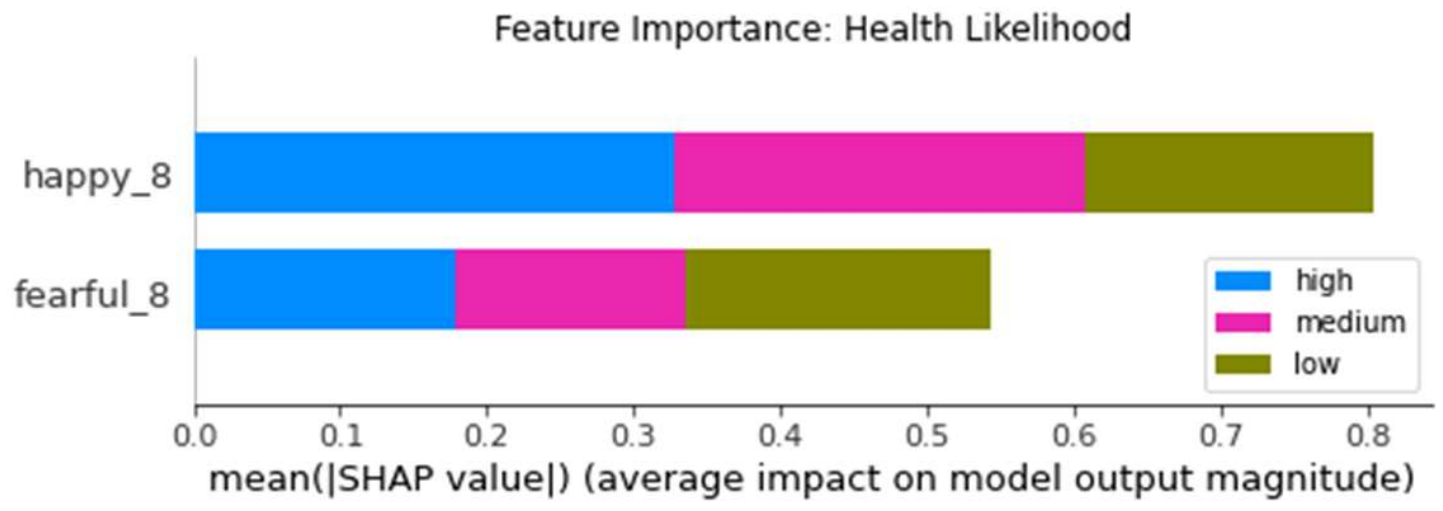

**Figure 6.** Feature importance for predicting health likelihood.

As the Shapley charts illustrate, again the emotions happiness and fear were found to have the strongest predictive power. However, we cannot make any claim about what emotional response to which movie predicts what personality characteristics. This is not the point of this paper. The point is that "your emotional response predicts your personality characteristics and moral values". Identification of the most emotionally provocative movies is most likely dependent on the individual personality and values of the viewer, which is also related to local cultures and values. It would therefore be another research project to precisely identify a minimal set of short movies that consistently provoke the most expressive emotions that are the most indicative of an individual's personality and morals.

## 5. Limitations, Future Work, and Conclusions

In this work, we show that AI can be used for the task of facial emotion recognition, producing features that can in turn predict people's personality and moral values.

Ours is an exploratory analysis with regard to associations found between different individual traits and emotions produced in response to a different set of audiovisual stimuli. These relationships could be further investigated in future research in order to better understand their meaning from a psychological perspective. Future research should consider more control variables, which we could not collect in our experiment (due to privacy arrangements), such as age, gender, and ethnicity of experiment participants. Similarly, a different set of videos could be taken into account, also looking for the optimal set of stimuli that could produce an emotional response better associable to specific individual differences.

Our research has both practical and theoretical implications. On the theoretical side, it further confirms the insight that moral affect—emotions in response to positive and negative experiences—are at the center of our ethical values. On the practical side, our approach offers a novel and more honest way to measure personality characteristics, attitudes to risk, and moral values. As has been discussed above, while humans tend to misjudge personality and moral values of others and themselves, AI provides an honest virtual mirror assisting in this task. In conclusion, this study has shown that while humans frequently are incapable of looking behind the facade of the face and "read the mind in the eyes", artificial intelligence can lend a helping hand to people who have difficulties in this task.

**Author Contributions:** Conceptualization, P.A.G.; methodology, P.A.G., A.F.C. and E.A.; software, C.C., M.F.K., L.R. and T.S.; formal analysis, P.A.G., A.F.C. and E.A.; investigation, C.C., M.F.K., L.R. and T.S.; data curation, C.C., M.F.K., L.R. and T.S.; writing—original draft preparation, P.A.G. and A.F.C.; writing—review and editing, P.A.G. and A.F.C. All authors have read and agreed to the published version of the manuscript.

**Funding:** This research received no external funding.

**Institutional Review Board Statement:** The study was conducted according to the guidelines of the Declaration of Helsinki, and approved by the Institutional Review Board of MIT (protocol code 170181783) on 16 February 2017.

**Informed Consent Statement:** Informed consent was obtained from all subjects involved in the study.

**Data Availability Statement:** The data presented in this study are available on request from the corresponding author. The data are not publicly available because they contain information that could compromise the privacy of research participants.

**Conflicts of Interest:** The authors declare no conflict of interest.

## Appendix A

Table A1 shows the Pearson's correlation coefficients of individual traits with the different emotions experienced while watching the videos. Each emotion is indicated together with the number of the video it is referring to. It is interesting to notice how significant associations emerge for every individual trait—with some emotions being particularly relevant for some traits, such as fear (revealed while watching videos 4–13 and 15) for the DOSPERT scale.

In the following, we provide additional charts that show the SHAP values of the features used for machine learning predictions (Figures A1–A18). We excluded those already presented in the results section.

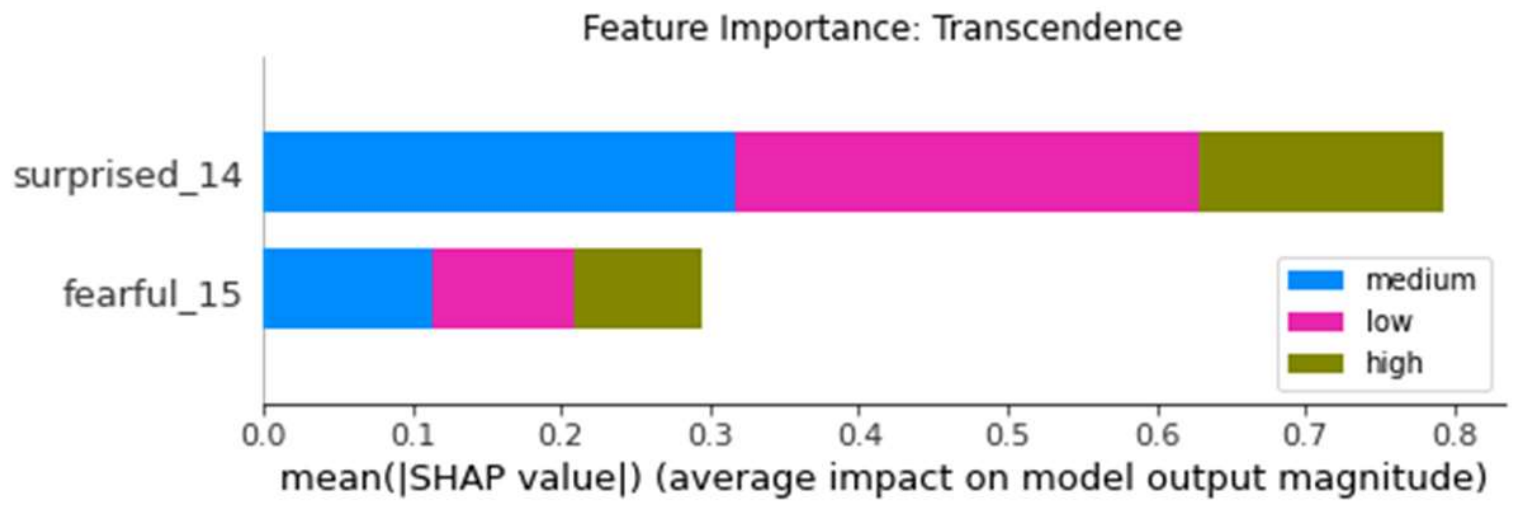

**Figure A1.** Feature importance for predicting transcendence.

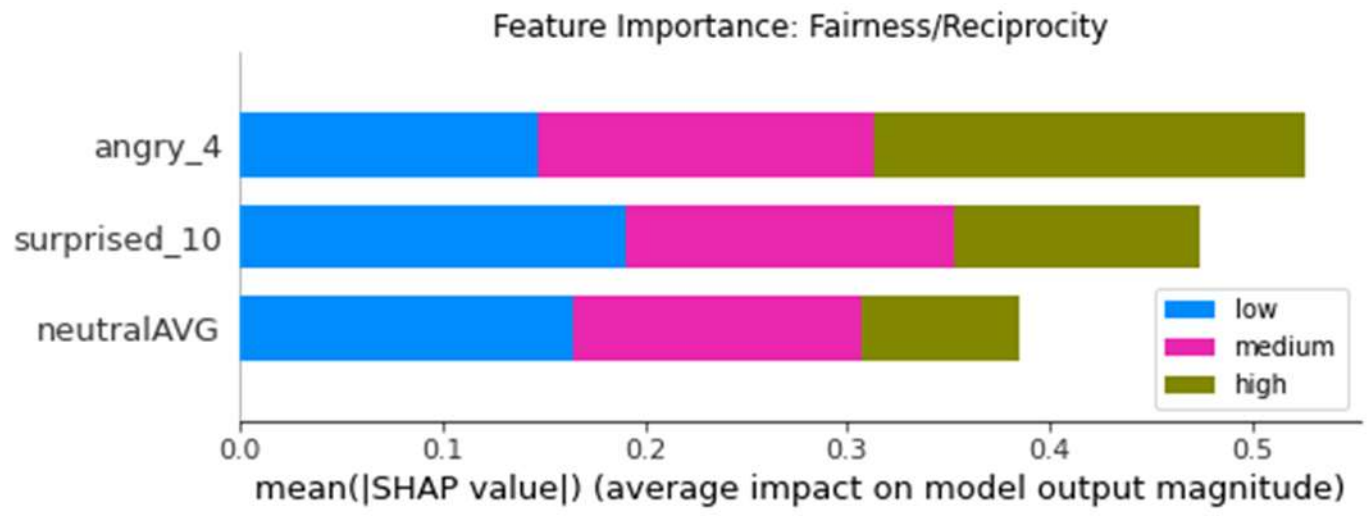

**Figure A2.** Feature importance for predicting fairness/reciprocity.

**Table A1.** Pearson's correlation coefficients.

| | Big Five | | | | | DOSPERT Scale | | | | | | | | | | Schwartz Values | | | Haidt Moral Values | | | |
|---|---|---|---|---|---|---|---|---|---|---|---|---|---|---|---|---|---|---|---|---|---|---|
| Variable | A | C | N | E | O | ETH_L | ETH_P | FIN_L | FIN_P | HEA_L | HEA_P | SOC_L | SOC_P | REC_L | REC_P | CON | TRA | HAR | FAIR | ING_LOY | AUTH | PUR |
| Angry 1 | −0.003 | 0.063 | 0.188 | 0.027 | 0.206 | −0.064 | −0.037 | −0.056 | 0.035 | −0.088 | 0.006 | 0.008 | 0.100 | −0.065 | 0.142 | 0.054 | 0.049 | 0.056 | 0.108 | 0.034 | 0.111 | −0.061 |
| Angry 2 | 0.104 | 0.156 | 0.233 * | 0.115 | 0.322 ** | −0.099 | 0.040 | −0.046 | 0.092 | −0.168 | 0.060 | −0.051 | 0.181 | −0.023 | 0.165 | 0.062 | 0.079 | 0.107 | 0.228 | 0.129 | 0.138 | −0.015 |
| Angry 3 | 0.118 | 0.091 | 0.192 | 0.191 | 0.141 | −0.020 | 0.013 | −0.025 | 0.064 | −0.094 | 0.075 | −0.015 | 0.117 | −0.014 | 0.088 | 0.031 | 0.006 | 0.101 | 0.187 | 0.069 | 0.097 | −0.055 |
| Angry 4 | 0.060 | 0.194 | 0.079 | 0.089 | 0.171 | 0.074 | 0.276 * | 0.088 | 0.136 | 0.097 | 0.190 | 0.199 | 0.322 ** | −0.052 | 0.199 | 0.172 | 0.094 | 0.178 | 0.296 * | 0.278 * | 0.299 * | 0.169 |
| Angry 5 | −0.078 | −0.034 | 0.152 | −0.011 | 0.263 * | 0.289 * | 0.044 | 0.173 | −0.042 | 0.047 | −0.010 | 0.138 | 0.250 * | −0.046 | 0.148 | −0.003 | −0.047 | 0.020 | −0.081 | −0.023 | −0.061 | −0.138 |
| Angry 6 | 0.035 | 0.048 | 0.181 | 0.039 | 0.203 | 0.094 | −0.015 | 0.083 | 0.013 | −0.029 | −0.075 | 0.101 | 0.240 | 0.013 | 0.111 | 0.077 | −0.048 | 0.045 | 0.115 | 0.023 | 0.056 | −0.025 |
| Angry 7 | 0.096 | 0.078 | 0.275 * | 0.100 | 0.277 * | −0.024 | 0.000 | 0.015 | 0.030 | −0.097 | 0.056 | −0.016 | 0.127 | −0.040 | 0.117 | 0.044 | 0.011 | 0.098 | 0.130 | 0.041 | 0.077 | −0.017 |
| Angry 8 | 0.065 | 0.070 | 0.235 * | 0.113 | 0.164 | −0.010 | 0.038 | 0.070 | 0.019 | −0.105 | 0.067 | −0.058 | 0.116 | −0.125 | 0.125 | 0.009 | 0.053 | 0.092 | 0.134 | 0.034 | 0.077 | 0.006 |
| Angry 9 | 0.004 | 0.062 | 0.211 | 0.069 | 0.248 * | 0.034 | −0.019 | 0.029 | 0.072 | −0.041 | 0.052 | −0.061 | 0.119 | −0.077 | 0.092 | 0.022 | −0.008 | 0.013 | 0.094 | 0.035 | 0.053 | −0.119 |
| Angry 10 | 0.048 | 0.089 | 0.243 * | 0.078 | 0.277 * | 0.003 | −0.001 | 0.013 | 0.080 | −0.048 | 0.054 | −0.013 | 0.112 | −0.027 | 0.107 | 0.057 | −0.007 | 0.038 | 0.098 | 0.028 | 0.062 | −0.079 |
| Angry 11 | −0.005 | 0.082 | 0.198 | 0.061 | 0.229 * | −0.024 | −0.014 | −0.022 | 0.058 | −0.078 | 0.057 | −0.031 | 0.121 | −0.034 | 0.103 | 0.070 | −0.004 | 0.055 | 0.115 | 0.028 | 0.084 | −0.061 |
| Angry 12 | 0.017 | 0.065 | 0.197 | 0.064 | 0.151 | −0.027 | −0.029 | 0.025 | 0.015 | −0.066 | −0.014 | −0.064 | 0.123 | −0.027 | 0.077 | 0.068 | 0.039 | 0.048 | 0.175 | 0.041 | 0.106 | −0.037 |
| Angry 13 | 0.103 | 0.049 | 0.142 | 0.171 | 0.088 | −0.083 | −0.029 | −0.007 | 0.011 | −0.109 | 0.037 | −0.085 | 0.147 | 0.009 | 0.157 | 0.071 | 0.016 | 0.192 | 0.225 | 0.064 | 0.115 | 0.038 |
| Angry 14 | −0.012 | −0.107 | 0.075 | 0.045 | 0.040 | −0.062 | −0.095 | 0.026 | −0.091 | −0.095 | −0.084 | −0.050 | 0.088 | −0.082 | 0.083 | −0.032 | 0.086 | 0.166 | 0.228 | 0.011 | −0.041 | −0.016 |
| Angry 15 | 0.099 | 0.000 | 0.060 | 0.220 | −0.005 | −0.097 | 0.013 | 0.091 | −0.029 | −0.111 | 0.063 | −0.093 | 0.160 | 0.024 | 0.183 | 0.034 | −0.024 | 0.209 | 0.169 | 0.108 | 0.114 | 0.082 |
| Disgusted 1 | 0.209 | 0.117 | 0.066 | 0.212 | −0.006 | −0.086 | 0.112 | 0.060 | 0.043 | −0.115 | 0.184 | −0.114 | 0.152 | 0.102 | 0.161 | 0.151 | −0.009 | 0.066 | 0.014 | 0.116 | 0.183 | 0.086 |
| Disgusted 2 | 0.234 * | 0.189 | 0.064 | 0.285 * | 0.087 | −0.114 | 0.088 | 0.041 | 0.025 | −0.126 | 0.140 | −0.040 | 0.241 | 0.168 | 0.174 | 0.088 | −0.063 | 0.177 | 0.129 | 0.169 | 0.178 | 0.146 |
| Disgusted 3 | 0.223 * | 0.190 | 0.073 | 0.295 ** | 0.044 | −0.125 | 0.073 | −0.013 | 0.059 | −0.059 | 0.158 | −0.128 | 0.175 | 0.116 | 0.175 | 0.135 | −0.045 | 0.119 | 0.148 | 0.139 | 0.219 | 0.140 |
| Disgusted 4 | 0.237 * | 0.185 | 0.058 | 0.330 ** | 0.046 | −0.120 | 0.110 | 0.033 | 0.045 | −0.100 | 0.174 | −0.073 | 0.200 | 0.148 | 0.184 | 0.105 | −0.056 | 0.158 | 0.118 | 0.160 | 0.200 | 0.161 |
| Disgusted 5 | 0.198 | 0.116 | 0.153 | 0.233 * | 0.071 | −0.122 | −0.064 | −0.094 | −0.019 | −0.059 | 0.019 | −0.087 | 0.099 | 0.043 | 0.107 | 0.069 | −0.094 | 0.043 | 0.058 | −0.024 | 0.118 | −0.015 |
| Disgusted 6 | 0.194 | 0.168 | 0.118 | 0.265 * | 0.122 | −0.091 | 0.059 | 0.015 | 0.054 | −0.057 | 0.163 | −0.104 | 0.186 | 0.062 | 0.176 | 0.101 | −0.040 | 0.129 | 0.182 | 0.134 | 0.187 | 0.078 |
| Disgusted 7 | 0.205 | 0.193 | 0.161 | 0.258 * | 0.073 | −0.110 | 0.027 | −0.054 | 0.069 | −0.032 | 0.144 | −0.168 | 0.131 | 0.044 | 0.143 | 0.142 | −0.035 | 0.076 | 0.190 | 0.102 | 0.218 | 0.080 |
| Disgusted 8 | 0.166 | 0.091 | 0.163 | 0.197 | −0.025 | −0.060 | −0.054 | −0.129 | 0.044 | 0.084 | 0.050 | −0.197 | −0.001 | −0.038 | 0.032 | 0.104 | −0.003 | −0.018 | 0.191 | −0.018 | 0.161 | 0.022 |
| Disgusted 9 | 0.105 | 0.105 | 0.204 | 0.203 | 0.155 | −0.049 | 0.010 | −0.001 | 0.036 | −0.092 | 0.074 | −0.037 | 0.143 | −0.018 | 0.122 | 0.081 | −0.020 | 0.084 | 0.107 | 0.047 | 0.117 | −0.014 |
| Disgusted 10 | 0.190 | 0.202 | 0.102 | 0.262 * | 0.114 | −0.108 | 0.109 | −0.004 | 0.094 | −0.079 | 0.194 | −0.109 | 0.199 | 0.088 | 0.211 | 0.161 | −0.054 | 0.099 | 0.135 | 0.140 | 0.212 | 0.088 |
| Disgusted 11 | 0.279 * | 0.177 | 0.134 | 0.225 * | 0.095 | −0.071 | 0.079 | −0.049 | 0.117 | −0.038 | 0.111 | −0.114 | 0.096 | 0.062 | 0.248 * | 0.043 | −0.204 | 0.170 | 0.180 | 0.041 | 0.211 | 0.138 |
| Disgusted 12 | 0.200 | 0.193 | 0.118 | 0.291 ** | 0.119 | −0.106 | 0.115 | 0.043 | 0.055 | −0.127 | 0.198 | −0.078 | 0.235 | 0.125 | 0.193 | 0.107 | −0.058 | 0.174 | 0.131 | 0.150 | 0.206 | 0.116 |
| Disgusted 13 | 0.211 | 0.147 | 0.035 | 0.253 * | −0.003 | −0.111 | 0.071 | −0.028 | 0.072 | −0.034 | 0.173 | −0.190 | 0.167 | 0.115 | 0.142 | 0.131 | −0.052 | 0.100 | 0.159 | 0.141 | 0.219 | 0.121 |
| Disgusted 14 | 0.221 * | 0.185 | 0.025 | 0.287 ** | 0.052 | −0.103 | 0.115 | 0.052 | 0.031 | −0.100 | 0.188 | −0.077 | 0.207 | 0.158 | 0.173 | 0.099 | −0.072 | 0.155 | 0.109 | 0.174 | 0.194 | 0.154 |
| Disgusted 15 | 0.213 | 0.163 | 0.070 | 0.314 ** | 0.106 | −0.117 | 0.169 | 0.035 | 0.063 | −0.148 | 0.219 | −0.015 | 0.175 | 0.148 | 0.204 | 0.051 | −0.162 | 0.200 | 0.090 | 0.149 | 0.159 | 0.148 |
| Surprised 1 | 0.064 | 0.117 | 0.010 | 0.053 | −0.072 | −0.056 | −0.007 | 0.036 | −0.019 | −0.114 | −0.045 | 0.108 | 0.229 | 0.186 | 0.253 * | 0.148 | −0.109 | 0.099 | 0.068 | 0.192 | 0.178 | 0.167 |
| Surprised 2 | −0.093 | −0.002 | 0.040 | −0.003 | −0.166 | −0.159 | 0.015 | −0.125 | 0.080 | −0.208 | 0.089 | 0.017 | −0.060 | 0.144 | 0.167 | 0.002 | 0.130 | 0.209 | 0.090 | 0.032 | −0.060 | 0.065 |
| Surprised 3 | 0.153 | 0.073 | 0.140 | 0.172 | −0.190 | −0.090 | −0.068 | −0.169 | −0.034 | 0.095 | −0.068 | 0.111 | −0.071 | 0.084 | 0.017 | 0.203 | −0.168 | 0.216 | 0.162 | 0.038 | 0.208 | 0.198 |
| Surprised 4 | −0.072 | −0.084 | 0.088 | 0.140 | −0.166 | 0.046 | −0.086 | −0.044 | −0.169 | 0.166 | −0.180 | 0.108 | −0.046 | −0.071 | −0.041 | 0.062 | −0.021 | 0.096 | 0.079 | −0.028 | 0.040 | 0.065 |
| Surprised 5 | 0.018 | 0.014 | 0.071 | 0.172 | 0.005 | 0.261 * | 0.088 | 0.155 | 0.167 | 0.283 * | 0.023 | 0.353 ** | 0.148 | 0.149 | 0.146 | 0.275 * | −0.162 | 0.106 | 0.074 | 0.179 | 0.097 | 0.129 |
| Surprised 6 | −0.157 | −0.110 | −0.090 | 0.032 | −0.122 | −0.039 | −0.325 ** | −0.032 | −0.151 | −0.018 | −0.236 | 0.094 | 0.030 | −0.011 | 0.138 | −0.095 | −0.012 | 0.157 | 0.011 | −0.001 | −0.217 | −0.141 |
| Surprised 7 | 0.011 | 0.076 | −0.004 | 0.131 | −0.283 * | −0.138 | −0.018 | −0.054 | −0.026 | −0.174 | 0.089 | −0.024 | −0.015 | 0.231 | −0.078 | 0.135 | −0.022 | 0.155 | 0.129 | 0.172 | 0.166 | 0.166 |
| Surprised 8 | −0.063 | −0.171 | 0.063 | 0.051 | −0.286 * | 0.060 | −0.074 | 0.036 | −0.111 | −0.111 | 0.051 | −0.006 | −0.145 | −0.151 | −0.025 | 0.076 | 0.170 | 0.083 | −0.063 | 0.150 | 0.120 | 0.120 |
| Surprised 9 | −0.068 | −0.008 | 0.233 * | 0.254 * | −0.027 | 0.034 | −0.024 | 0.038 | −0.123 | 0.058 | −0.079 | −0.034 | −0.062 | −0.030 | −0.090 | 0.096 | −0.008 | 0.011 | −0.137 | −0.013 | 0.052 | −0.041 |
| Surprised 10 | −0.114 | −0.266 * | 0.054 | −0.024 | −0.216 | 0.085 | −0.087 | 0.089 | −0.202 | −0.077 | −0.093 | −0.014 | −0.168 | −0.139 | −0.052 | 0.142 | 0.190 | 0.103 | −0.185 | −0.097 | −0.092 | −0.055 |
| Surprised 11 | −0.034 | −0.099 | 0.112 | 0.014 | 0.142 | 0.081 | 0.195 | 0.174 | 0.143 | −0.023 | 0.245 * | 0.099 | −0.014 | 0.163 | 0.151 | 0.222 | −0.168 | 0.220 | 0.004 | 0.202 | 0.015 | 0.094 |
| Surprised 12 | 0.025 | −0.009 | −0.013 | 0.033 | −0.135 | 0.062 | 0.042 | −0.005 | −0.072 | 0.072 | −0.035 | 0.263 * | 0.042 | 0.175 | −0.089 | 0.114 | 0.058 | 0.062 | 0.133 | 0.161 | 0.167 | 0.139 |
| Surprised 13 | 0.053 | −0.026 | 0.081 | 0.122 | −0.132 | 0.050 | 0.058 | 0.088 | −0.057 | −0.172 | 0.165 | −0.038 | 0.075 | 0.098 | 0.082 | 0.186 | 0.088 | 0.176 | −0.062 | 0.174 | 0.213 | 0.205 |
| Surprised 14 | 0.011 | −0.025 | 0.115 | −0.236 * | 0.086 | −0.050 | −0.046 | 0.013 | 0.018 | −0.091 | 0.016 | −0.053 | −0.122 | −0.023 | 0.079 | 0.310 ** | −0.209 | 0.089 | −0.093 | 0.144 | 0.188 | 0.271 * |
| Surprised 15 | 0.095 | 0.067 | 0.231 * | 0.293 ** | −0.069 | −0.004 | −0.021 | −0.024 | −0.079 | 0.045 | −0.021 | 0.038 | −0.018 | 0.158 | −0.054 | 0.166 | −0.131 | 0.144 | −0.001 | 0.178 | 0.257 * | 0.190 |
| Fearful 1 | −0.112 | −0.033 | 0.050 | −0.072 | −0.132 | −0.066 | 0.017 | −0.095 | −0.061 | −0.215 | 0.109 | 0.004 | −0.170 | 0.067 | 0.032 | −0.056 | 0.063 | −0.018 | −0.106 | −0.058 | −0.132 | −0.014 |
| Fearful 2 | −0.083 | −0.017 | 0.057 | −0.047 | −0.127 | −0.143 | 0.144 | 0.037 | 0.047 | −0.201 | 0.212 | −0.025 | 0.198 | 0.155 | 0.221 | 0.083 | −0.032 | 0.174 | 0.093 | 0.183 | 0.178 | 0.185 |
| Fearful 3 | −0.150 | −0.039 | 0.115 | −0.012 | −0.164 | −0.074 | 0.043 | −0.052 | −0.179 | −0.191 | 0.130 | −0.004 | −0.224 | 0.087 | −0.136 | −0.150 | −0.145 | 0.146 | 0.018 | −0.096 | −0.125 | 0.092 |
| Fearful 4 | 0.049 | 0.143 | −0.058 | 0.021 | 0.074 | 0.088 | 0.279 * | −0.146 | 0.138 | 0.221 | 0.207 | 0.466 ** | 0.122 | 0.029 | 0.113 | 0.086 | 0.063 | 0.160 | 0.226 | 0.186 | 0.173 | 0.199 |
| Fearful 5 | −0.018 | 0.045 | −0.024 | 0.022 | 0.055 | 0.179 | 0.336 ** | −0.023 | 0.218 | 0.315 * | 0.251 * | 0.406 ** | 0.099 | 0.053 | 0.111 | 0.135 | 0.039 | 0.154 | 0.154 | 0.128 | 0.109 | 0.108 |
| Fearful 6 | −0.119 | −0.050 | 0.039 | −0.055 | −0.133 | −0.129 | −0.024 | −0.195 | −0.040 | −0.041 | 0.080 | 0.029 | −0.002 | 0.030 | 0.184 | 0.036 | −0.076 | 0.172 | 0.145 | −0.067 | −0.067 | 0.016 |
| Fearful 7 | −0.108 | −0.021 | 0.051 | −0.037 | −0.085 | 0.330 ** | 0.369 ** | 0.169 | 0.378 ** | 0.386 ** | 0.353 ** | 0.389 ** | 0.181 | 0.225 | 0.250 * | 0.130 | 0.033 | 0.163 | 0.170 | 0.258 * | 0.100 | 0.118 |
| Fearful 8 | −0.112 | −0.030 | 0.053 | −0.072 | −0.129 | −0.079 | −0.021 | −0.196 | 0.001 | −0.220 | 0.112 | 0.029 | −0.030 | −0.085 | 0.103 | 0.024 | −0.009 | 0.003 | 0.032 | 0.041 | 0.086 | 0.116 |
| Fearful 9 | −0.182 | −0.085 | 0.056 | −0.010 | −0.033 | 0.350 ** | 0.204 | 0.427 ** | 0.375 ** | 0.214 | 0.285 * | 0.066 | 0.105 | 0.265 * | 0.336 ** | 0.092 | −0.057 | 0.006 | −0.096 | 0.180 | −0.026 | −0.068 |
| Fearful 10 | −0.149 | −0.053 | 0.046 | −0.061 | −0.091 | 0.262 * | 0.208 | 0.322 ** | 0.404 ** | 0.080 | 0.335 ** | 0.057 | 0.076 | 0.251 * | 0.371 ** | 0.063 | 0.034 | 0.009 | −0.014 | 0.148 | −0.043 | −0.095 |
| Fearful 11 | 0.112 | 0.062 | 0.113 | 0.065 | −0.105 | 0.213 | 0.231 | 0.109 | 0.217 | 0.097 | 0.193 | 0.123 | 0.033 | 0.177 | 0.269 * | 0.045 | −0.030 | 0.099 | 0.086 | 0.035 | −0.009 | 0.104 |
| Fearful 12 | 0.150 | 0.054 | 0.097 | 0.084 | −0.084 | 0.143 | 0.152 | 0.154 | 0.229 | 0.026 | 0.307 * | 0.216 | 0.071 | 0.124 | 0.359 ** | 0.111 | 0.047 | 0.108 | −0.025 | 0.166 | 0.091 | 0.109 |
| Fearful 13 | −0.030 | 0.005 | −0.085 | −0.114 | −0.103 | 0.018 | −0.051 | −0.021 | 0.138 | −0.058 | 0.084 | 0.090 | −0.032 | 0.025 | 0.386 ** | 0.076 | 0.000 | 0.047 | −0.079 | 0.049 | 0.005 | 0.050 |
| Fearful 14 | 0.061 | −0.009 | 0.037 | −0.154 | 0.021 | 0.014 | 0.003 | −0.117 | 0.095 | 0.024 | −0.097 | −0.002 | −0.159 | −0.079 | 0.201 | −0.118 | −0.272 * | 0.102 | 0.046 | −0.149 | 0.093 | 0.130 |
| Fearful 15 | −0.076 | −0.149 | 0.138 | 0.034 | 0.114 | 0.272 * | −0.010 | 0.059 | −0.110 | 0.165 | 0.007 | −0.046 | 0.109 | 0.098 | 0.065 | −0.119 | −0.229 | −0.011 | −0.154 | 0.018 | −0.086 | −0.025 |
| Sad 1 | −0.003 | 0.129 | 0.176 | 0.130 | −0.001 | −0.038 | 0.106 | −0.119 | 0.103 | −0.102 | 0.228 | −0.142 | 0.056 | 0.013 | 0.142 | 0.069 | 0.060 | 0.186 | 0.215 | 0.237 * | 0.207 | 0.298 * |
| Sad 2 | 0.102 | 0.118 | 0.153 | 0.135 | −0.035 | −0.051 | 0.080 | 0.012 | −0.053 | −0.023 | 0.191 | −0.144 | 0.151 | 0.041 | 0.138 | 0.095 | −0.020 | 0.207 | 0.130 | 0.212 | 0.265 * | 0.363 ** |
| Sad 3 | −0.096 | 0.007 | 0.070 | 0.112 | 0.028 | 0.043 | 0.180 | 0.066 | 0.096 | 0.010 | 0.229 | −0.172 | 0.205 | −0.008 | 0.232 | 0.179 | 0.160 | 0.170 | 0.213 | 0.285 * | 0.256 * | 0.284 * |
| Sad 4 | 0.031 | 0.036 | 0.150 | 0.031 | 0.045 | 0.028 | 0.058 | −0.119 | 0.020 | 0.058 | 0.158 | −0.106 | 0.036 | −0.038 | 0.114 | −0.031 | −0.136 | 0.179 | 0.137 | 0.160 | 0.214 | 0.327 ** |
| Sad 5 | −0.021 | −0.013 | 0.137 | 0.076 | 0.040 | 0.083 | 0.110 | −0.260 * | 0.231 | 0.127 | 0.156 | −0.039 | 0.012 | −0.043 | 0.216 | 0.149 | −0.079 | 0.177 | 0.238 * | 0.080 | 0.136 | 0.164 |
| Sad 6 | −0.055 | 0.009 | 0.038 | −0.038 | −0.030 | 0.032 | 0.110 | −0.219 | 0.095 | 0.071 | 0.090 | 0.062 | 0.009 | −0.096 | 0.164 | −0.037 | 0.117 | 0.240 * | 0.348 ** | 0.103 | 0.145 | 0.295 * |

**Table A1.** *Cont.*

| | | | Big Five | | | | | | | DOSPERT Scale | | | | | | Schwartz Values | | | | Haidt Moral Values | | |
|---|---|---|---|---|---|---|---|---|---|---|---|---|---|---|---|---|---|---|---|---|---|---|
| Variable | A | C | N | E | O | ETH_L | ETH_P | FIN_L | FIN_P | HEA_L | HEA_P | SOC_L | SOC_P | REC_L | REC_P | CON | TRA | HAR | FAIR | ING_LOY | AUTH | PUR |
| Sad 7 | −0.075 | 0.012 | 0.033 | 0.017 | −0.006 | 0.133 | 0.127 | −0.014 | 0.149 | 0.165 | 0.153 | 0.042 | 0.047 | 0.014 | 0.192 | 0.029 | 0.119 | 0.175 | 0.279 * | 0.208 | 0.141 | 0.260 * |
| Sad 8 | −0.039 | 0.046 | 0.079 | −0.008 | 0.039 | 0.156 | 0.132 | 0.000 | 0.165 | 0.217 | 0.119 | 0.045 | 0.029 | 0.033 | 0.176 | 0.078 | 0.059 | 0.167 | 0.235 | 0.205 | 0.255 * | 0.290 * |
| Sad 9 | −0.062 | 0.002 | 0.101 | −0.046 | 0.002 | 0.072 | 0.031 | −0.059 | 0.062 | 0.108 | 0.066 | −0.096 | 0.008 | −0.035 | 0.147 | 0.041 | 0.043 | 0.147 | 0.141 | 0.161 | 0.244 * | 0.330 ** |
| Sad 10 | −0.013 | 0.021 | 0.068 | 0.007 | 0.037 | 0.066 | 0.056 | −0.064 | 0.101 | 0.176 | 0.087 | −0.103 | 0.020 | −0.052 | 0.188 | 0.098 | 0.014 | 0.128 | 0.197 | 0.202 | 0.259 * | 0.309 ** |
| Sad 11 | −0.030 | −0.032 | 0.036 | 0.105 | −0.049 | −0.030 | 0.089 | −0.222 | 0.173 | 0.045 | 0.230 | −0.106 | 0.114 | −0.173 | 0.190 | −0.062 | 0.018 | 0.227 | 0.326 ** | 0.204 | 0.119 | 0.156 |
| Sad 12 | −0.126 | −0.125 | 0.035 | 0.058 | −0.101 | 0.091 | 0.062 | −0.131 | 0.152 | 0.133 | 0.182 | −0.023 | 0.046 | −0.097 | 0.213 | −0.064 | −0.056 | 0.259 * | 0.306 * | 0.187 | 0.070 | 0.153 |
| Sad 13 | −0.114 | −0.139 | 0.039 | −0.062 | −0.024 | 0.189 | 0.086 | −0.049 | 0.194 | 0.172 | 0.139 | −0.042 | 0.097 | −0.098 | 0.200 | −0.004 | 0.094 | 0.237 * | 0.299 * | 0.213 | 0.150 | 0.197 |
| Sad 14 | −0.047 | −0.084 | 0.159 | 0.042 | −0.017 | 0.150 | 0.049 | −0.094 | 0.171 | 0.082 | 0.156 | −0.056 | 0.033 | −0.082 | 0.127 | 0.020 | −0.062 | 0.190 | 0.257 * | 0.183 | 0.160 | 0.176 |
| Sad 15 | −0.124 | −0.112 | 0.077 | −0.033 | 0.023 | 0.134 | 0.028 | −0.037 | 0.193 | 0.092 | 0.133 | −0.210 | 0.095 | −0.120 | 0.215 | 0.023 | 0.073 | 0.140 | 0.226 | 0.168 | 0.178 | 0.116 |
| Happy 1 | −0.195 | −0.297 ** | 0.014 | 0.091 | −0.220 * | 0.078 | −0.107 | −0.041 | 0.072 | 0.031 | 0.138 | −0.126 | −0.170 | −0.048 | 0.135 | −0.007 | −0.075 | 0.343 ** | 0.132 | −0.085 | −0.134 | −0.035 |
| Happy 2 | −0.191 | −0.193 | −0.019 | 0.131 | −0.145 | 0.065 | −0.009 | 0.011 | 0.068 | 0.040 | 0.091 | 0.131 | −0.135 | 0.044 | 0.078 | −0.143 | −0.012 | 0.415 ** | 0.223 | −0.056 | −0.223 | −0.034 |
| Happy 3 | −0.187 | −0.099 | 0.005 | 0.095 | −0.187 | −0.092 | −0.060 | −0.005 | −0.088 | −0.040 | −0.045 | 0.072 | −0.152 | −0.006 | −0.021 | −0.272 * | −0.113 | 0.205 | 0.052 | −0.143 | −0.182 | 0.021 |
| Happy 4 | −0.168 | −0.053 | 0.115 | 0.024 | −0.105 | −0.032 | −0.152 | 0.056 | −0.099 | −0.017 | −0.025 | 0.039 | 0.012 | −0.164 | 0.013 | −0.195 | 0.179 | 0.203 | 0.081 | 0.005 | −0.076 | 0.033 |
| Happy 5 | −0.107 | −0.211 | 0.111 | −0.009 | −0.121 | −0.018 | −0.221 | 0.069 | −0.060 | −0.024 | −0.040 | −0.014 | −0.086 | −0.124 | −0.194 | −0.320 ** | 0.101 | 0.187 | 0.068 | −0.025 | −0.104 | −0.059 |
| Happy 6 | −0.100 | −0.012 | 0.056 | 0.030 | 0.037 | −0.018 | −0.116 | 0.059 | 0.150 | 0.018 | 0.056 | 0.024 | 0.107 | −0.142 | 0.080 | −0.270 * | −0.025 | 0.081 | 0.071 | 0.046 | −0.152 | −0.175 |
| Happy 7 | 0.004 | 0.121 | 0.080 | 0.061 | 0.015 | −0.092 | −0.126 | −0.027 | −0.010 | −0.037 | 0.066 | −0.005 | 0.143 | −0.053 | 0.014 | −0.185 | −0.098 | 0.098 | 0.077 | 0.139 | −0.027 | −0.026 |
| Happy 8 | 0.035 | 0.121 | 0.155 | 0.272 * | 0.111 | −0.196 | 0.021 | −0.092 | 0.078 | −0.257 * | 0.273 * | 0.003 | 0.134 | 0.017 | 0.116 | −0.163 | −0.290 * | 0.206 | 0.036 | 0.141 | −0.136 | −0.079 |
| Happy 9 | −0.123 | −0.112 | 0.113 | 0.168 | 0.028 | −0.099 | −0.029 | 0.055 | −0.096 | −0.120 | 0.141 | 0.013 | 0.035 | 0.003 | 0.025 | −0.327 ** | −0.247 * | 0.147 | 0.005 | 0.058 | −0.235 | −0.071 |
| Happy 10 | −0.194 | −0.093 | −0.004 | 0.134 | 0.044 | −0.065 | −0.027 | 0.015 | −0.103 | −0.092 | 0.108 | −0.007 | 0.028 | 0.006 | 0.002 | −0.348 ** | −0.278 * | 0.061 | −0.103 | 0.010 | −0.319 ** | −0.177 |
| Happy 11 | −0.136 | −0.089 | 0.015 | 0.041 | 0.125 | 0.096 | −0.042 | 0.105 | −0.124 | 0.126 | −0.103 | −0.047 | 0.108 | 0.019 | −0.144 | −0.270 * | −0.183 | −0.016 | −0.194 | −0.079 | −0.115 | −0.105 |
| Happy 12 | −0.004 | −0.049 | 0.063 | 0.152 | 0.019 | 0.012 | 0.052 | 0.079 | −0.179 | −0.065 | 0.078 | 0.034 | 0.052 | 0.044 | −0.074 | −0.099 | −0.170 | 0.194 | −0.062 | −0.039 | −0.080 | 0.003 |
| Happy 13 | −0.164 | 0.009 | −0.080 | 0.088 | 0.044 | −0.096 | −0.138 | −0.056 | 0.092 | −0.127 | 0.127 | 0.159 | 0.216 | −0.228 | 0.163 | −0.201 | 0.020 | 0.141 | 0.081 | 0.124 | −0.207 | −0.228 |
| Happy 14 | −0.081 | −0.165 | −0.017 | 0.023 | −0.180 | 0.034 | −0.127 | 0.001 | −0.091 | 0.000 | 0.034 | 0.007 | 0.037 | −0.298 * | 0.106 | −0.196 | 0.083 | 0.160 | 0.074 | −0.008 | −0.220 | −0.136 |
| Happy 15 | −0.113 | −0.061 | 0.283 * | 0.141 | 0.038 | −0.108 | −0.121 | −0.109 | −0.084 | −0.094 | −0.009 | 0.085 | −0.022 | −0.001 | −0.078 | −0.268 * | −0.205 | 0.175 | 0.007 | −0.002 | −0.223 | −0.042 |
| Neutral 1 | 0.197 | 0.239 * | −0.085 | −0.125 | 0.227 * | −0.055 | 0.067 | 0.076 | −0.102 | 0.015 | −0.208 | 0.163 | 0.129 | 0.034 | −0.192 | −0.026 | 0.056 | −0.396 ** | −0.201 | −0.005 | 0.048 | −0.067 |
| Neutral 2 | 0.159 | 0.147 | −0.040 | −0.176 | 0.164 | −0.038 | −0.021 | −0.013 | −0.053 | −0.020 | −0.158 | −0.080 | 0.073 | −0.066 | −0.133 | 0.107 | 0.017 | −0.483 ** | −0.268 * | −0.019 | 0.129 | −0.086 |
| Neutral 3 | 0.180 | 0.069 | −0.047 | −0.165 | 0.165 | 0.087 | −0.011 | −0.011 | 0.044 | 0.042 | −0.053 | 0.002 | 0.054 | −0.006 | −0.078 | 0.179 | 0.062 | −0.270 * | −0.143 | 0.024 | 0.057 | −0.132 |
| Neutral 4 | 0.070 | −0.022 | −0.210 | −0.126 | 0.053 | 0.021 | 0.037 | 0.036 | 0.057 | −0.031 | −0.127 | 0.033 | −0.095 | 0.126 | −0.140 | 0.138 | −0.046 | −0.335 ** | −0.208 | −0.159 | −0.146 | −0.295 * |
| Neutral 5 | 0.070 | 0.163 | −0.236 * | −0.103 | 0.036 | −0.075 | 0.105 | 0.127 | −0.132 | −0.097 | −0.094 | 0.008 | 0.010 | 0.126 | −0.040 | 0.143 | −0.001 | −0.315 ** | −0.246 * | −0.045 | −0.032 | −0.063 |
| Neutral 6 | 0.075 | −0.031 | −0.103 | −0.041 | −0.015 | −0.005 | −0.034 | 0.164 | −0.153 | −0.053 | −0.118 | −0.062 | −0.129 | 0.134 | −0.251 * | 0.127 | −0.078 | −0.301 * | −0.401 ** | −0.146 | −0.095 | −0.192 |
| Neutral 7 | 0.038 | −0.110 | −0.130 | −0.091 | −0.031 | −0.055 | −0.054 | 0.029 | −0.146 | −0.115 | −0.205 | −0.021 | −0.154 | 0.006 | −0.212 | 0.057 | −0.043 | −0.237 * | −0.331 ** | −0.289 * | −0.146 | −0.224 |
| Neutral 8 | −0.012 | −0.130 | −0.228 * | −0.228 * | −0.084 | 0.028 | −0.109 | 0.078 | −0.186 | 0.023 | −0.304 * | −0.009 | −0.119 | −0.017 | −0.231 | 0.038 | 0.148 | −0.287 * | −0.234 | −0.272 * | −0.133 | −0.180 |
| Neutral 9 | 0.159 | 0.076 | −0.255 * | −0.172 | −0.068 | −0.003 | −0.012 | −0.039 | 0.003 | 0.009 | −0.188 | 0.072 | −0.066 | 0.017 | −0.170 | 0.193 | 0.169 | −0.237 * | −0.108 | −0.185 | −0.022 | −0.163 |
| Neutral 10 | 0.171 | 0.078 | −0.099 | −0.138 | −0.067 | −0.009 | −0.020 | 0.015 | 0.014 | −0.038 | −0.169 | 0.100 | −0.044 | 0.052 | −0.180 | 0.150 | 0.181 | −0.180 | −0.064 | −0.166 | 0.041 | −0.084 |
| Neutral 11 | 0.053 | 0.045 | −0.112 | −0.166 | −0.077 | −0.017 | −0.104 | 0.140 | −0.153 | −0.081 | −0.226 | 0.125 | −0.186 | 0.119 | −0.188 | 0.129 | 0.118 | −0.263 * | −0.257 * | −0.188 | −0.103 | −0.122 |
| Neutral 12 | 0.073 | 0.094 | −0.096 | −0.143 | 0.082 | −0.077 | −0.080 | 0.111 | −0.139 | −0.103 | −0.212 | 0.024 | −0.097 | 0.069 | −0.235 | 0.046 | 0.071 | −0.304 * | −0.336 ** | −0.212 | −0.107 | −0.169 |
| Neutral 13 | 0.106 | 0.112 | −0.042 | −0.008 | 0.014 | −0.146 | −0.068 | 0.058 | −0.213 | −0.121 | −0.192 | 0.046 | −0.171 | 0.115 | −0.261 * | 0.007 | −0.094 | −0.291 * | −0.339 ** | −0.260 * | −0.158 | −0.180 |
| Neutral 14 | 0.030 | 0.106 | −0.157 | −0.086 | 0.058 | −0.129 | −0.020 | 0.082 | −0.137 | −0.052 | −0.184 | 0.069 | −0.077 | 0.151 | −0.201 | 0.023 | 0.057 | −0.273 * | −0.297 * | −0.199 | −0.123 | −0.162 |
| Neutral 15 | 0.145 | 0.120 | −0.264 * | −0.116 | −0.047 | −0.047 | 0.036 | 0.083 | −0.108 | −0.016 | −0.126 | 0.139 | −0.096 | 0.076 | −0.165 | 0.113 | 0.079 | −0.256 * | −0.213 | −0.175 | −0.067 | −0.106 |

* $p < 0.05$; ** $p < 0.01$. N = neuroticism; C = conscientiousness; A = agreeableness; O = openness to experience; E = extraversion; ETH_L = ethical likelihood; ETH_P = ethical perceived; FIN_L = financial likelihood; FIN_P = financial perceived; HEA_L = health likelihood; HEA_P = health perceived; SOC_L = social likelihood; SOC_P = social perceived; REC_L = recreational likelihood; REC_P = recreational perceived; CON = conservation; TRA = transcendence; HAR = harm/care; FAIR = fairness/reciprocity; ING_LOY = in-group loyalty; AUTH = authority/respect; PUR = purity/sanctity.

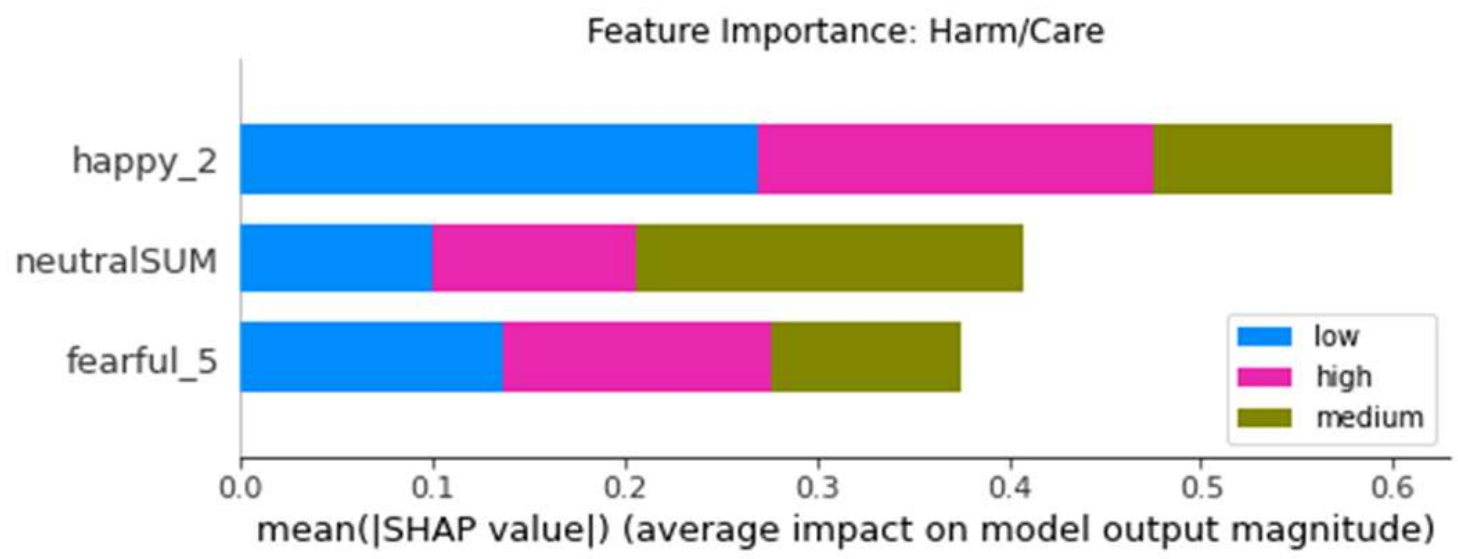

**Figure A3.** Feature importance for predicting harm/care.

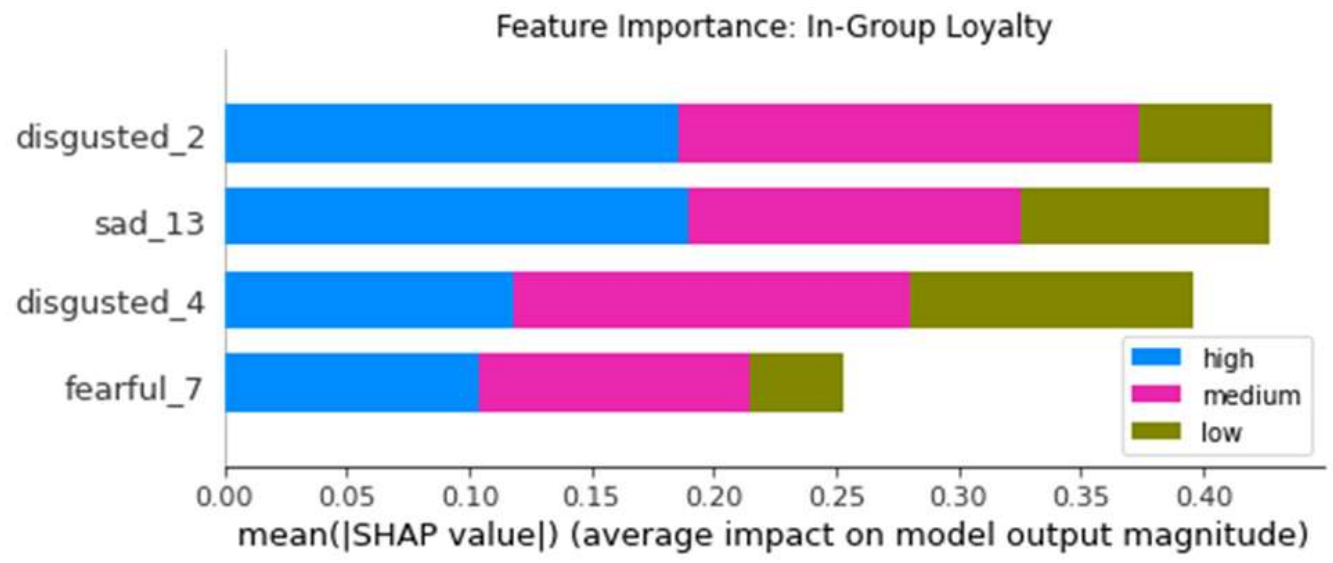

**Figure A4.** Feature importance for predicting in-group loyalty.

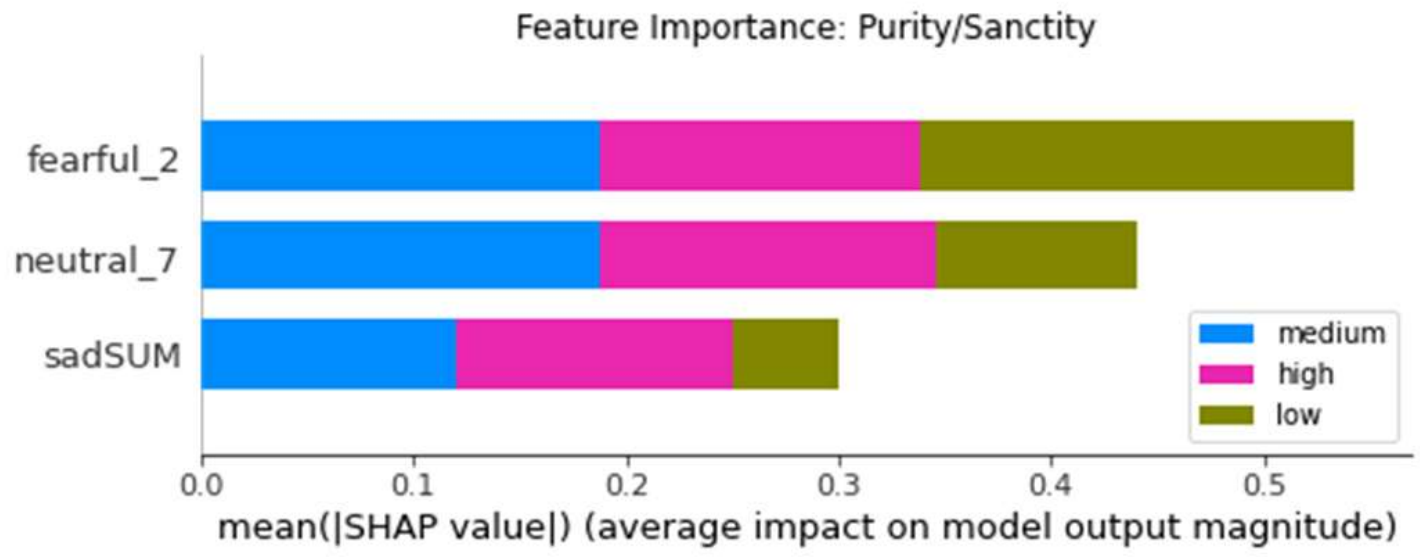

**Figure A5.** Feature importance for predicting purity/sanctity.

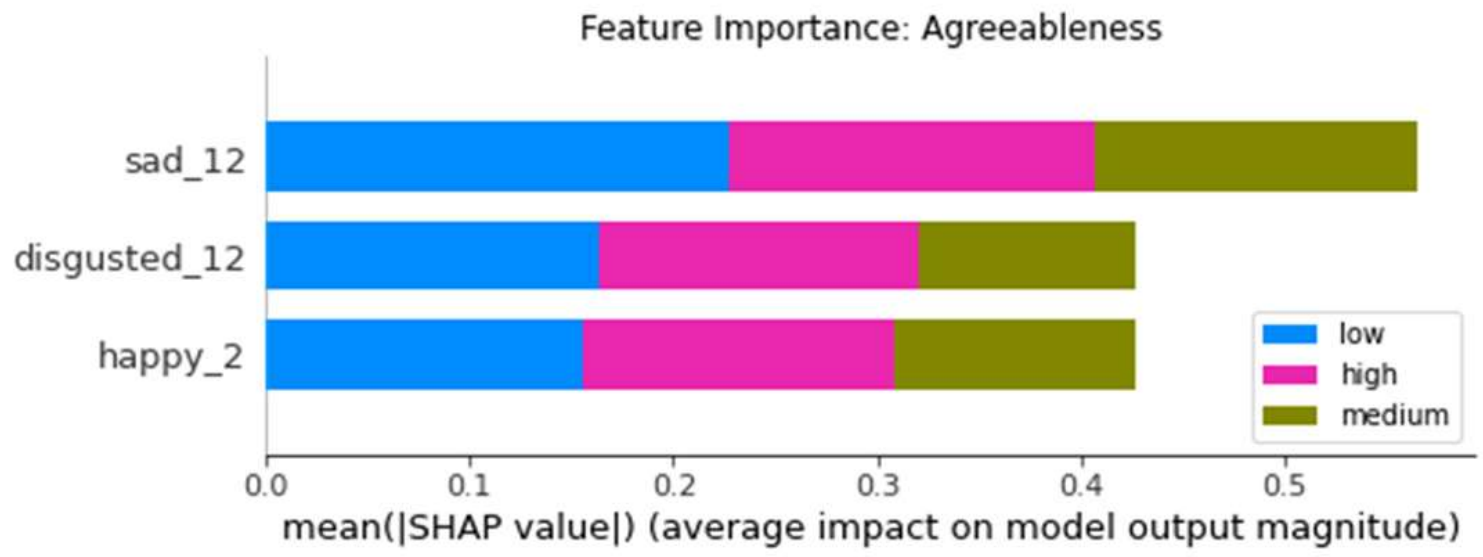

**Figure A6.** Feature importance for predicting agreeableness.

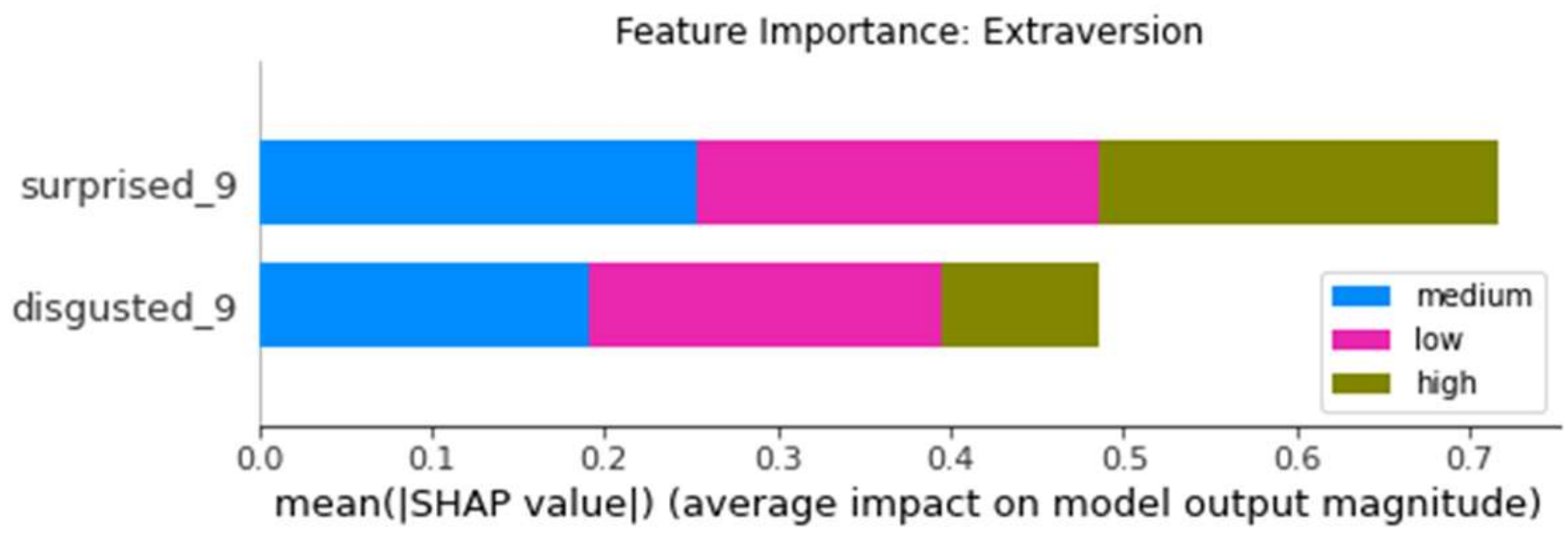

**Figure A7.** Feature importance for predicting extraversion.

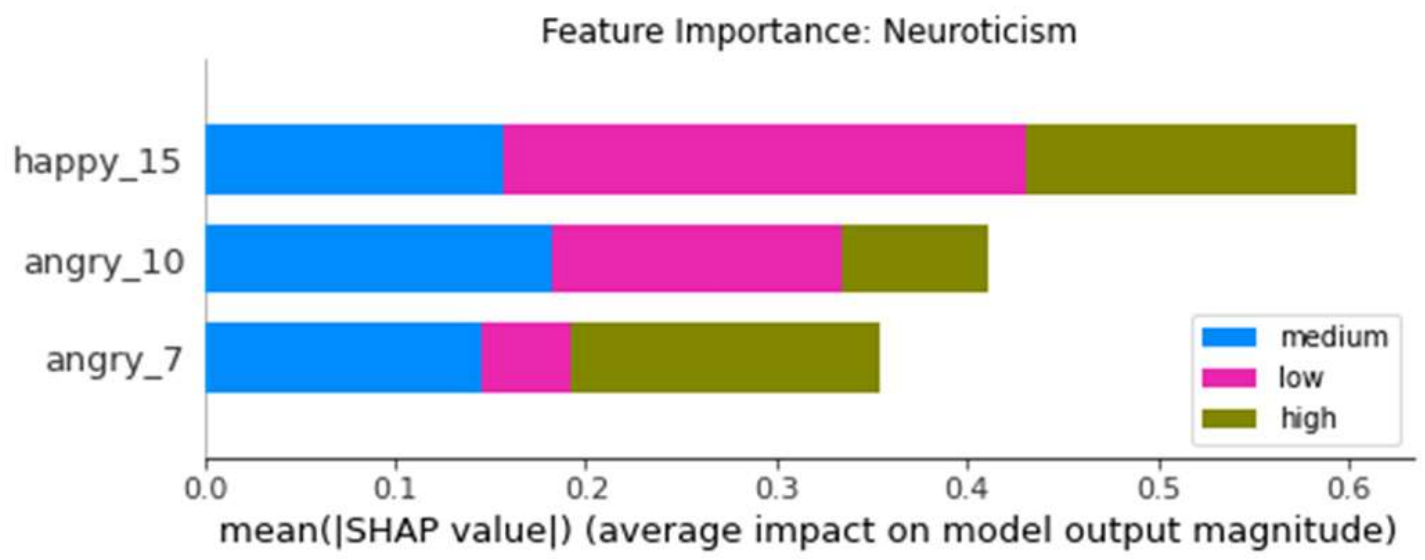

**Figure A8.** Feature importance for predicting neuroticism.

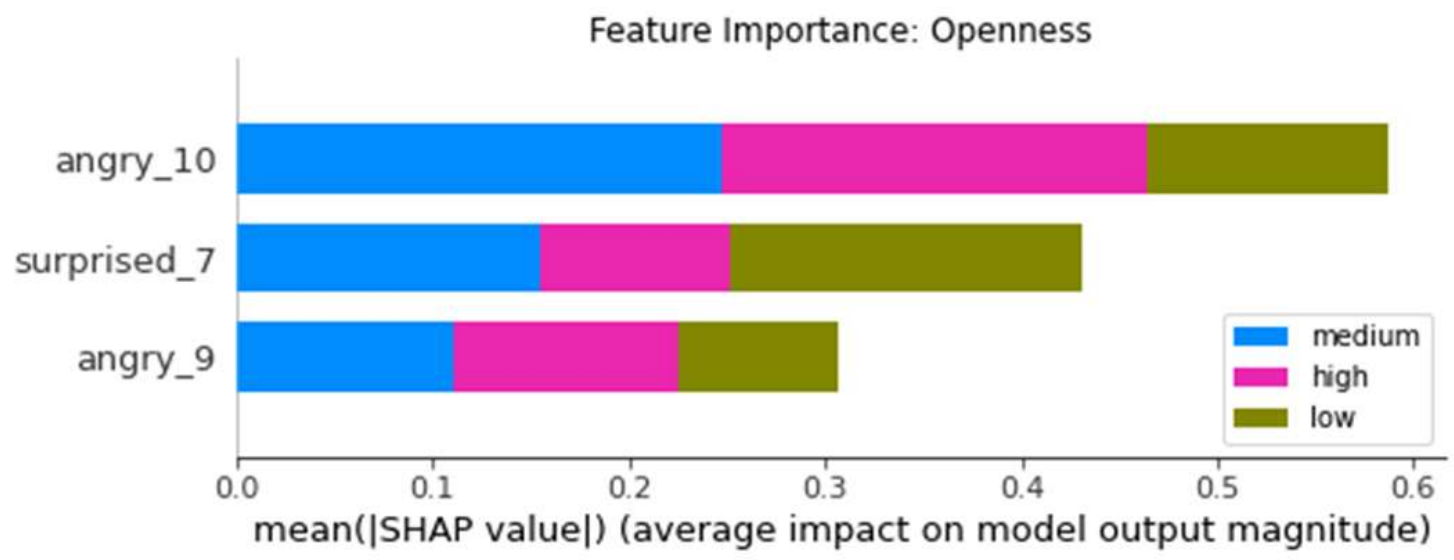

**Figure A9.** Feature importance for predicting openness.

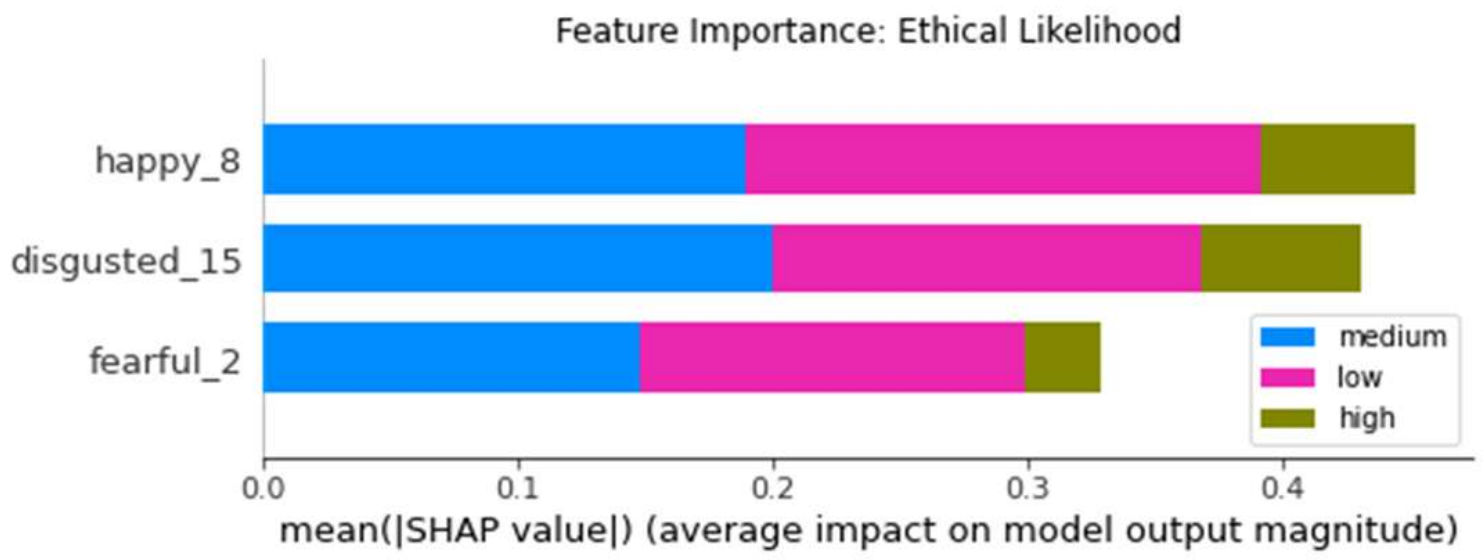

**Figure A10.** Feature importance for predicting ethical likelihood.

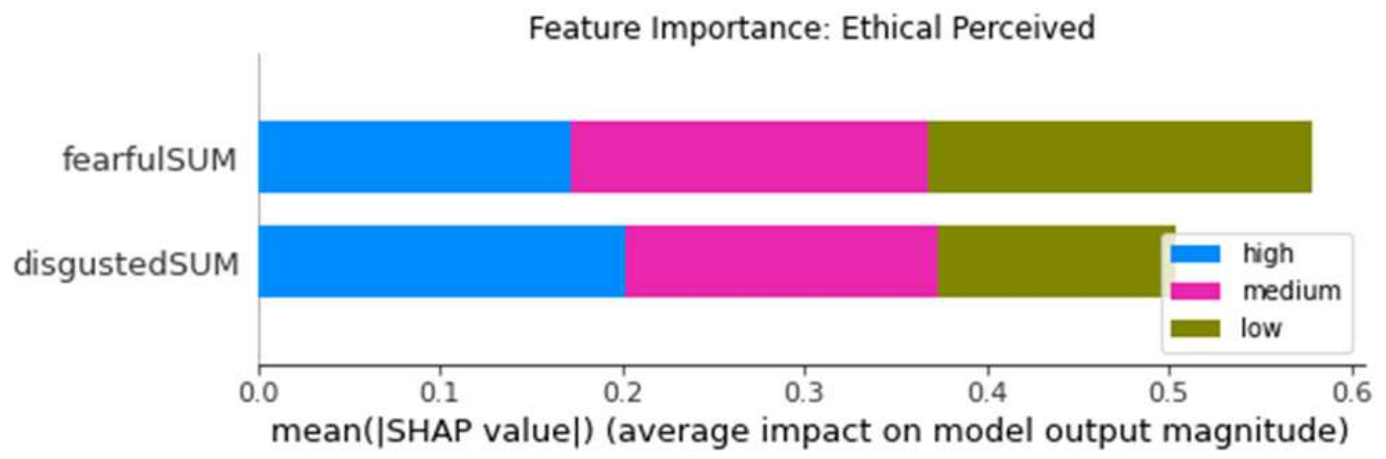

**Figure A11.** Feature importance for predicting ethical perceived.

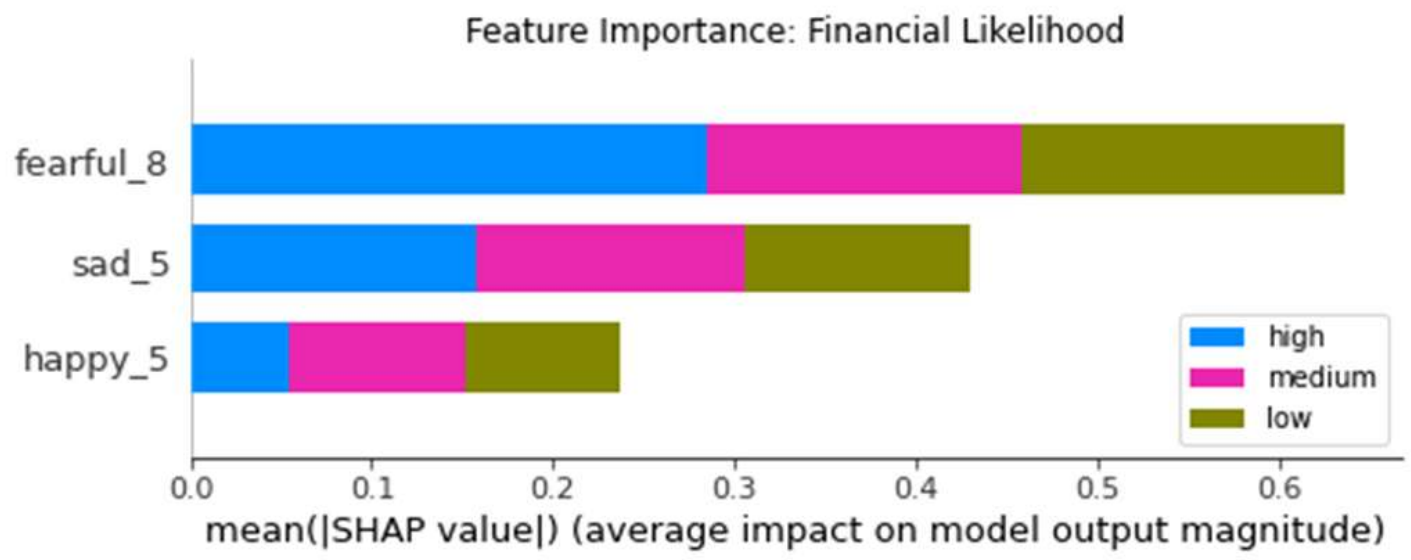

**Figure A12.** Feature importance for predicting financial likelihood.

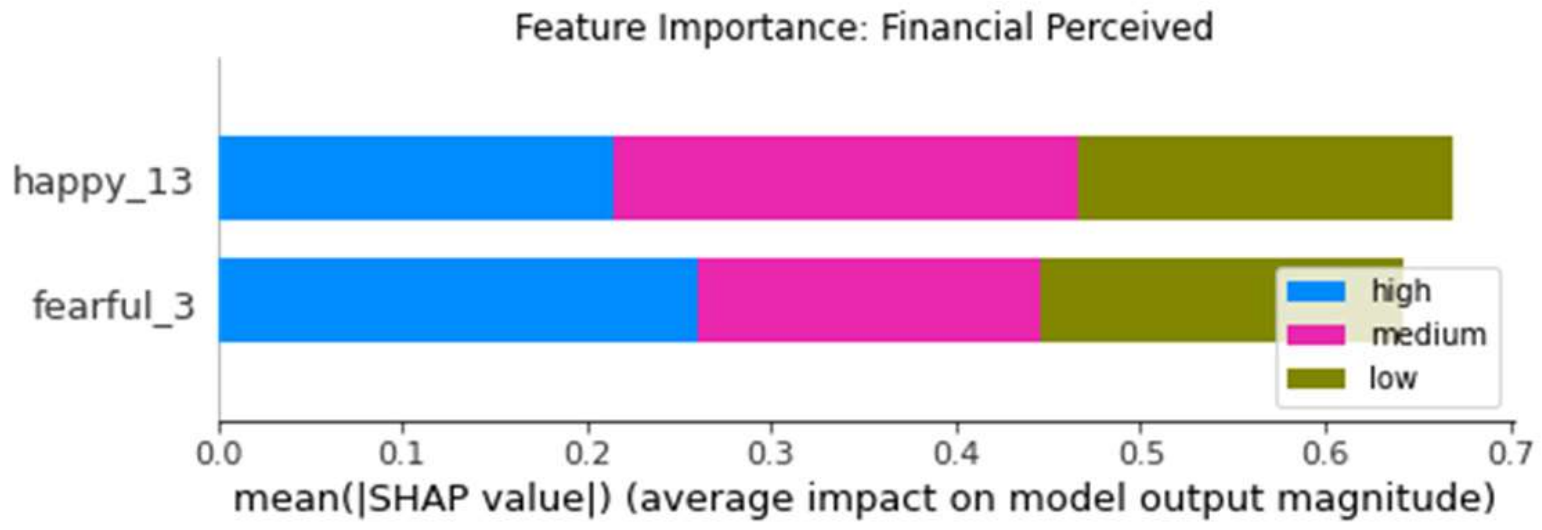

**Figure A13.** Feature importance for predicting financial perceived.

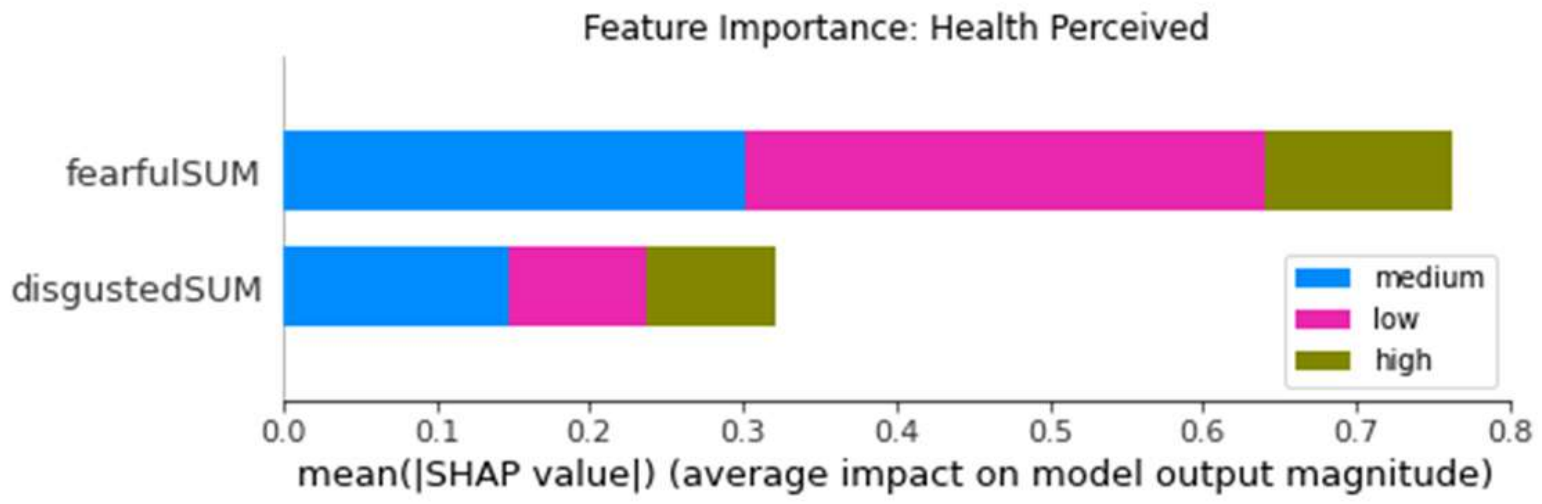

**Figure A14.** Feature importance for predicting health perceived.

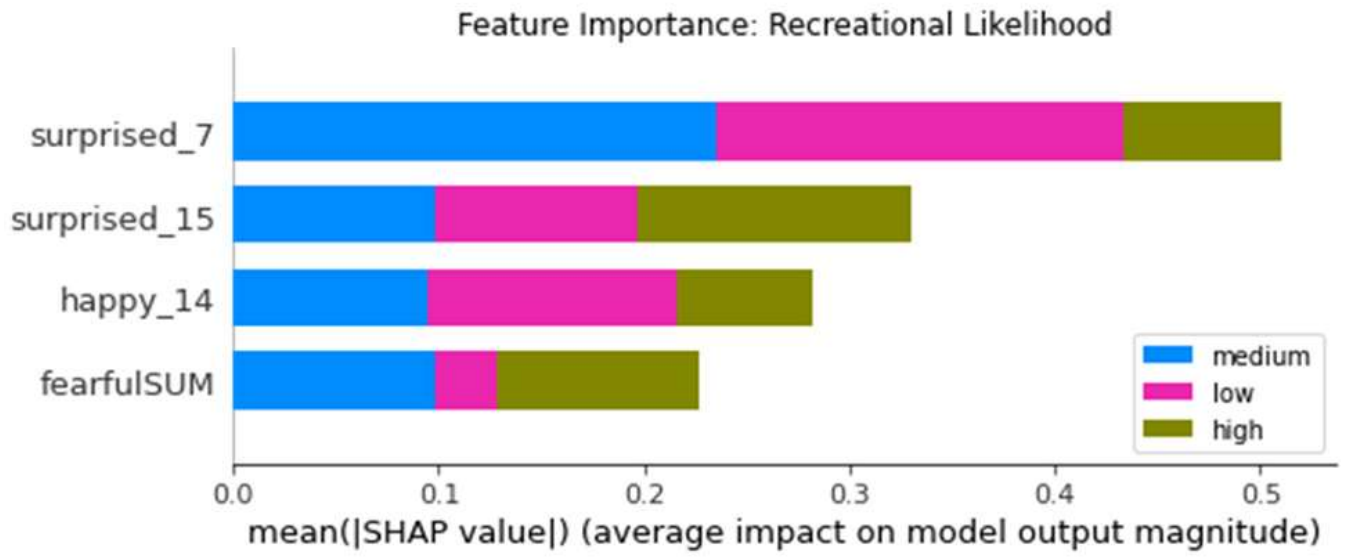

**Figure A15.** Feature importance for predicting recreational likelihood.

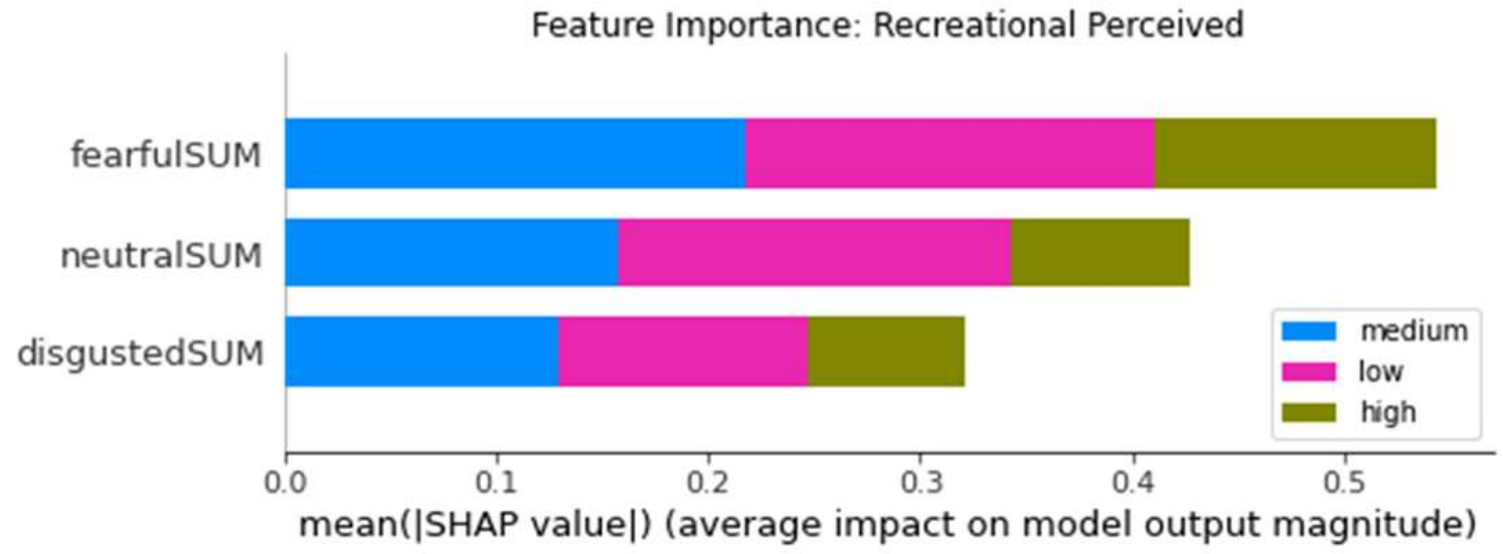

**Figure A16.** Feature importance for predicting recreational perceived.

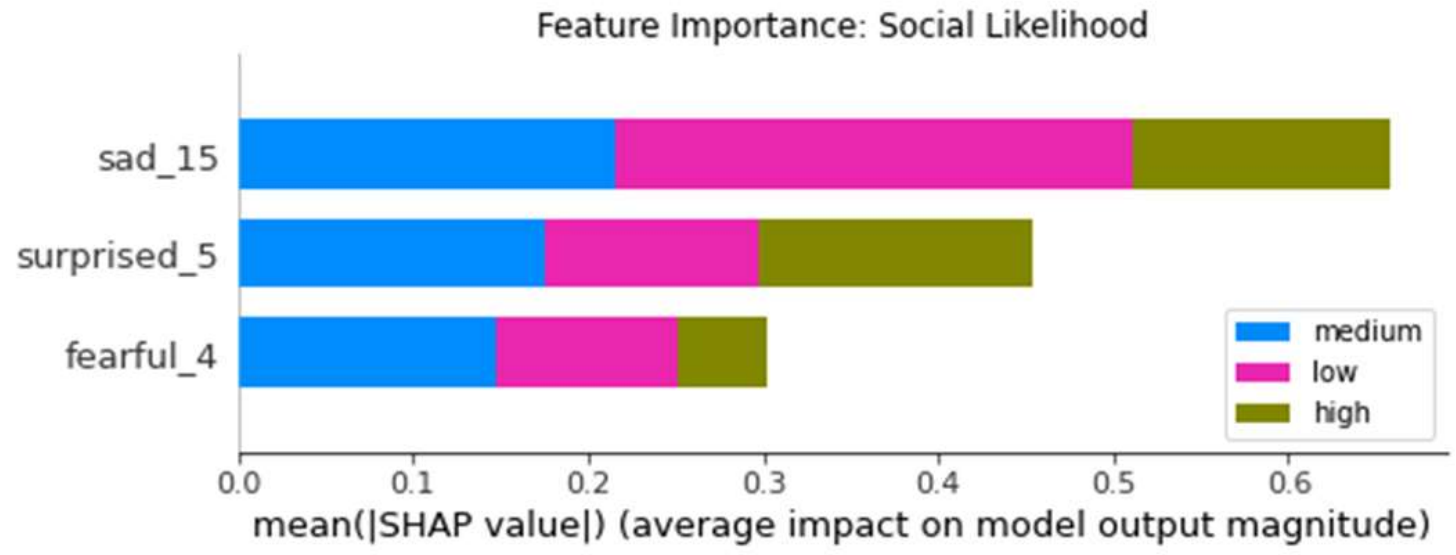

**Figure A17.** Feature importance for predicting social likelihood.

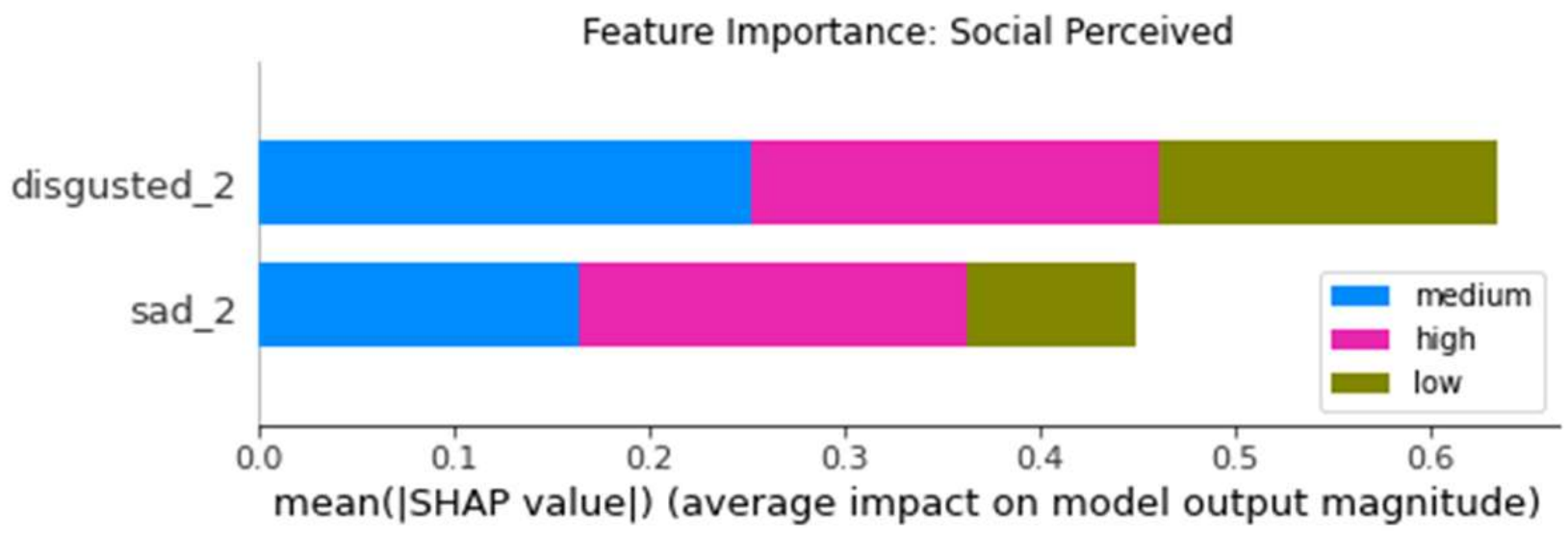

**Figure A18.** Feature importance for predicting social perceived.